\pdfoutput=1
\documentclass[11pt]{article}

\usepackage[]{EMNLP2022}

\usepackage{inconsolata}
\usepackage{times}
\usepackage{latexsym}
\usepackage{booktabs}
\usepackage{multirow}
\usepackage{graphicx}
\usepackage{hyperref}
\usepackage{paralist}
\usepackage{enumitem}

\usepackage[T1]{fontenc}
\usepackage[utf8]{inputenc}

\usepackage{microtype}

\usepackage{amsmath}
\usepackage{amssymb}
\usepackage{xspace}
\usepackage[ruled,vlined]{algorithm2e}
\newcommand{\ourmodel}{\textsc{MuCoLa}\xspace}
\long\def\ignore#1{}

\title{Gradient-Based Constrained Sampling from Language Models}
\author{Sachin Kumar$^\clubsuit$ \quad Biswajit Paria$^\heartsuit$ \quad Yulia Tsvetkov$^\spadesuit$ \\
$^\clubsuit$Language Technologies Institute, Carnegie Mellon University, Pittsburgh PA \\
$^\heartsuit$Machine Learning Department, Carnegie Mellon University, Pittsburgh PA \\
 $^\spadesuit$Paul G.~Allen School of Computer Science \& Engineering, University of Washington, Seattle WA \\
\texttt{\small \{sachink,bparia\}@cs.cmu.edu, yuliats@cs.washington.edu}}

\begin{document}
\maketitle

\newcommand{\Sref}[1]{\S\ref{#1}}
\newcommand{\fref}[1]{figure~\ref{#1}}
\newcommand{\Fref}[1]{Figure~\ref{#1}}
\newcommand{\tref}[1]{table~\ref{#1}}
\newcommand{\Tref}[1]{Table~\ref{#1}}
\newcommand{\Aref}[1]{Appendix~\ref{#1}}

\begin{abstract}
Large pretrained language models generate fluent text but are notoriously hard to controllably sample from. 
In this work, we study constrained sampling from such language models: generating text that satisfies user-defined constraints, while maintaining fluency and model's performance in a downstream task. 
%Standard autoregressive decoding strategies are not always conducive to imposing such constraints globally.
We propose \ourmodel---a sampling procedure that combines the log-likelihood of the language model with arbitrary (differentiable) constraints in a single energy function, and then generates samples in a non-autoregressive manner. 
Specifically, it initializes the entire output sequence with noise and follows a Markov chain defined by Langevin Dynamics using the gradients of the energy function. 
We evaluate \ourmodel on text generation with soft and hard constraints as well as their combinations 
% obtaining competitive results in 
%and their combinations and 
obtaining significant improvements over competitive baselines for toxicity avoidance, sentiment control, and keyword-guided generation.\footnote{The code is available at: \url{https://github.com/Sachin19/mucoco/tree/sampling}}

% Yet, it is difficult to  
% We present \ourmodel--a general purpose decoding algorithm from language models 
\end{abstract}

\begin{figure}
    \centering
    \includegraphics[width=0.45\textwidth]{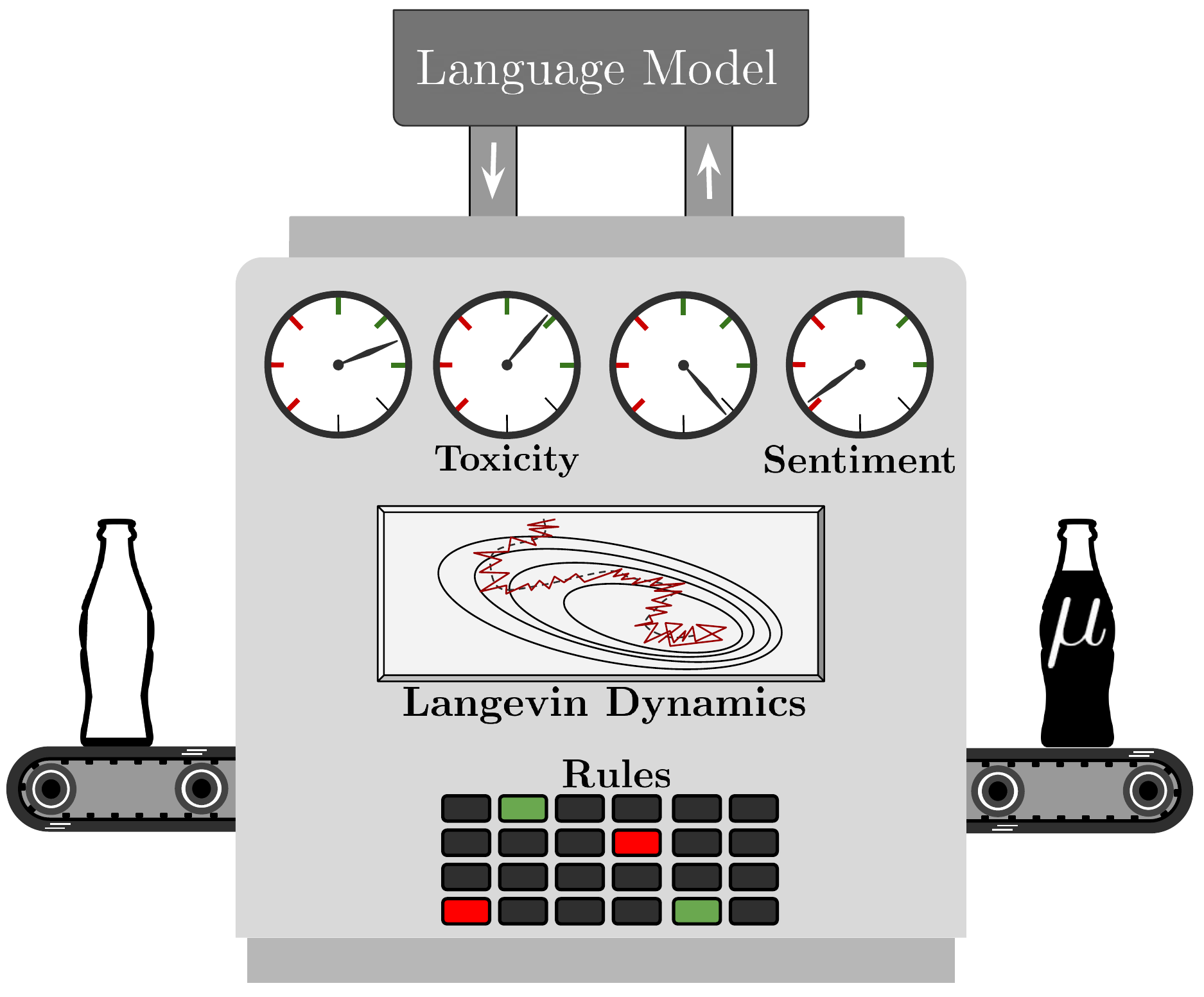}
    \caption{\ourmodel, our proposed method, stylized as $\mu$\textsc{CoLa}. Given a language model, a prompt/input $\mathbf{x}$, and desired constraints defined as thresholds on differentiable functions, we perform Langevin Dynamics updates to generate the entire output sequence $\mathbf{y}$ non-autogressively. We show experiments highlighting both hard and soft constraints (\Sref{sec:experiments}).}
    \label{fig:mucola}
\end{figure}

\section{Introduction}
\label{sec:introduction}

Transformer-based language models (LMs) %\citep{vaswani2017attention} 
trained on web-scale corpora \citep{radford2019language, 2020t5, gpt3} 
are generating impressively realistic texts. Despite having human-level fluency, they are far from reaching human-level communication abilities and are hard to control for content, context, and intent in communication.
This results in unreliable models that 
lack basic knowledge, 
hallucinate facts, and 
discriminate users~\citep{10.1145/3442188.3445922,gehman-etal-2020-realtoxicityprompts,pagnoni2021understanding}. 

Controlled text generation---sampling text from LMs to satisfy constraints on the properties of generated text---aims to address these issues.
Prior works incorporate constraints in existing decoding algorithms at token level by modifying output probabilities~\citep{Dathathri2020Plug,yang-klein-2021-fudge,KrauseGeDi2020,liu2021onthefly,lu-etal-2021-neurologic,pascual-etal-2021-plug-play,liu-etal-2021-dexperts}. While effective in certain settings, by generating autoregressively (i.e., left-to-right), these approaches fail to account for global context and hardly generalize beyond a single constraint. More importantly, by modifying output probabilities, they end up altering the underlying LM distribution compromising the fluency and task accuracy \cite{kumar2021controlled}. 
%For example, if the goal is generate a polite translation of a source sentence using a translation model, the output should be a sample from the model in that it should not forgo meaning in lieu of politeness. 

We propose an algorithm to sample text \emph{non-autoregressively} from a conditional or unconditional LM trained to perform any language generation task---translation, summarization, dialog, prompt completion---while controlling for multiple, potentially competing constraints, and without sacrificing the base model quality.
Combining the LM likelihood with constraints into a single ``energy'' function, our algorithm follows a Markov Chain~\citep{brooks2011handbook} to iteratively transform an output sequence initialized randomly into a desired output with low energy.

Since common Monte Carlo Markov Chain (MCMC) sampling methods can be intractably slow~\citep{Sokal1997}, we propose to define this Markov Chain using \emph{gradients of the energy function} with respect to \emph{token embeddings} of the output sequence. Additionally, we introduce stochasticity in the process in order to generate diverse samples, by modifying the gradients with additive noise, a process referred to as Langevin Dynamics ~\citep{10.2307/2346184,1981NuPhB.180..378P,10.5555/3104482.3104568,doi:10.1137/0329055,song-etal-2020-adversarial}. Finally, we operationalize the energy function by defining each constraint to be smaller than threshold, and writing it as a Lagrangian---with language model likelihood as the primary objective~\citep{kumar2021controlled}. 
Besides allowing us to combine any number of constraints of varying scales without a need of tuning their weights, we show that low-energy solutions under this definition are true samples from the LM distribution.
We call this algorithm \ourmodel for sampling with \textbf{mu}ltiple \textbf{co}nstraints from LMs using \textbf{La}ngevin Dynamics (\Sref{sec:method}; also see \fref{fig:mucola} in the Appendix).

We show the efficacy and generality of \ourmodel on a variety of tasks, LMs and constraints from prior work (\Sref{sec:experiments}), including soft constraints (\Sref{sec:soft-constraints}) defined using auxiliary models (e.g., classifiers or smaller LMs), as well as hard rule-based constraints (\Sref{sec:hard-constraints}) defined by presence or absence of certain keyphrases. We conduct experiments on (a) toxicity avoidance, sentiment control in GPT2-Large, and (b) keyword controlled generation with GPT2-Large and XL, and (c) entity control to improve fidelity in translation. Through both automatic metrics and human evaluation, we show versatility of this method through improved or matched performance with competitive baselines, in terms of quality and diversity. Finally, we present preliminary results on new tasks, %of entity-controlled summarization and combination of hard and soft constraints 
showing promising directions for future research. % [not sure of this last line].

% and summarization models and performance improvements on both automatic metrics and human evaluation.

\section{Background: Constrained Sampling from Language Models}
\label{sec:background}
Let $P(\mathbf{y}|\mathbf{x}; \theta)$ model the conditional probability distribution of an output token sequence $\mathbf{y} = (y_1, \ldots, y_N)$, given an optional input token sequence $\mathbf{x} = (x_1, \ldots, x_M)$ where $x_m, y_n \in \mathcal{V}$.%, the vocabulary. 

\ignore{Traditionally, the decoder consists of a input layer $\mathbf{E}$ which first converts each discrete $y_n$ to a dense vector $e_{y_n}$ via an embedding table lookup (also referred to as a non-contextual embedding). This vector is then fed through a series of neural network layers (e.g., transformers or LSTMs) to obtain a hidden state (or a contextual token embedding) after which an output embedding layer projects it back to vocabulary space using the softmax operation. To reduce number of trainable parameters, most modern text generation systems usually share the input and output embedding tables~\citep{press-wolf-2017-using}.

Given $\mathbf{x}$, decoding from such a model involves finding outputs $\mathbf{y} \in \mathcal{Y}$ which admit a high probability under $P$. Since searching through the space of all possible output sequences $\mathcal{Y}$ is intractable, most decoding algorithms factorize $P$ over each token $y_n$, where the output is generated left-to-right, with the output token in step $n$ being fed to the input at step $n+1$. It typically involves search or sampling strategies like beam search, ancestral sampling, top-k sampling~\citep{fan-etal-2018-hierarchical}, or nucleus sampling~\citep{Holtzman2020The}, among others.
}
%constrained sampling definition
We are interested in \emph{constrained sampling} from $P$---finding output sequences $\mathbf{y}$ that have a high probability under $P$ while minimizing a given set of constraint functions: $\{f_1, \ldots, f_C\}$. 
We assume that each
% $f_i$
% can be computed by a function 
$f_i:([\mathbf{x}], \mathbf{y}) \rightarrow \mathbb{R}$ is defined such that a lower value of $f_i$ implies that the output better satisfies the constraint.
% adheres to the control. 
For example, to constrain the outputs to only non-toxic continuations for a given prompt $\mathbf{x}$,
% if we want $\mathbf{y}$ to not be ``toxic'',
we define a classifier $p_\textsc{toxic} (\mathbf{y})$ which predicts the output toxicity probability, with lower probability implying lower toxicity.
We assume all $f_i$ are differentiable. 

Enforcing these constraints in an autoregressive (i.e., left-to-right) decoding strategy like beam search or sampling is challenging, since the constraints are defined conceptually on the whole output sequence and are hard to evaluate accurately only on the generated prefix~\citep{yang2021fudge, liu2021onthefly}. With multiple constraints, their satisfaction/balancing becomes challenging.
% computationally expensive. 
% [maybe talk about most of them work only for classifier/language model like constraints or just in related work]. 
%soft relaxation
%
\begin{figure*}
    \centering
    \includegraphics[width=\textwidth]{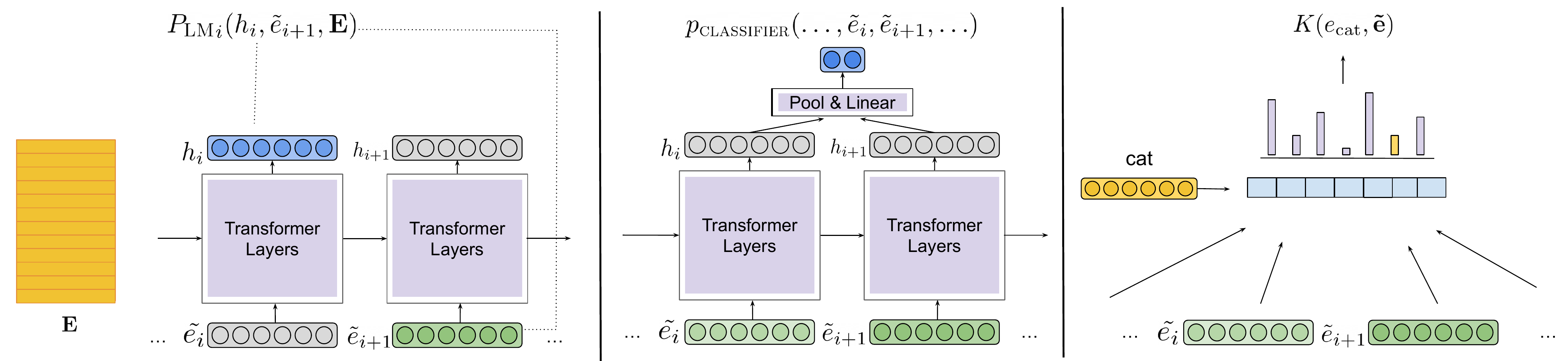}
    \caption{Different kinds of functions can be incorporated into \ourmodel defined on a shared embedding table $\mathbf{E}$. (Left) Language Modeling objective defines a per-token loss directly on the sequence of embeddings. For every token this loss provides gradients to update $\tilde{e}_i$ via backpropagation through the transformer layers and directly to $\tilde{e}_{i+1}$ through the negative loss likelihood loss as computed in \Sref{subsec:lagrangian}. This is used as a primary objective for the underlying LM and can also be used for classification as discussed in \Sref{subsec:sentiment-controlled-generation} (Center) Classification objective defined on probability of the desired label. The classifier gets the token embeddings $\tilde{\mathbf{e}}$ directly as input and updates the embedding using gradients obtained via backpropagation from the transformer layers (Right) Lexical loss defined on the embeddings directly (without the use of additional models) to include desired keywords or phrases in the output sequence (\Sref{sec:hard-constraints}). In practice any combination of these constraints can be used.}
    \label{fig:primary-objective}
\end{figure*}
%\paragraph{Non-autoregressive controlled generation} 
Recent work thus explored \textbf{non-autoregressive controlled generation} ~\citep{kumar2021controlled}, using constrained \textit{optimization} over $\mathbf{y}$---finding a single output $\mathbf{y}$ which \emph{maximizes} $P$ given the constraints by performing gradient descent on the outputs $\mathbf{y}$. 
This involves (1) representing the constrained optimization problem as a single objective $\mathcal{E}(\mathbf{y})$ (often referred to as an \emph{energy function}, discussed in \Sref{subsec:lagrangian}), and (2) relaxing the discrete outputs $\mathbf{y}$ to continuous approximations such that gradient descent is feasible. 
% First, the constrained optimization is converted to a single energy function (we discuss how this energy function $\mathcal{E}(\mathbf{y})$ is represented in \Sref{}). 
In previous works, the latter is achieved by creating a soft-representation of $\mathbf{y}$, $\tilde{\mathbf{y}} = (\tilde{y}_1, \ldots, \tilde{y}_N)$ where each $\tilde{y}_n \in \mathbb{R}^{|\mathcal{V}|}$ is a simplex (or ``logits'' which are converted to a simplex using softmax) over the target vocabulary $\mathcal{V}$, representing the probability of the $n$-th token in the sequence. We refer to these methods as gradient-based decoding.
Representing the decoding objective as $\min_\mathbf{\tilde{y}} \mathcal{E}(\mathbf{\tilde{y}})$ and initializing $\tilde{\mathbf{y}}$ with $\tilde{\mathbf{y}}^0$, it is updated as
\begin{equation}
    \tilde{\mathbf{y}}^t = \tilde{\mathbf{y}}^{t-1} - \eta \nabla_{\tilde{\mathbf{y}}} \mathcal{E}(\tilde{\mathbf{y}}^{t-1}),
    \label{eq:expgd}
\end{equation}
% \end{align}
where $\eta > 0$ denotes the step size. In this process, the underlying LMs (and functions $f_i$) remain fixed and are used to provide gradients to the sequence $\mathbf{\tilde{y}}$.
After performing multiple steps of this gradient descent,
% with the loss function $\mathcal{L}(x, \tilde{\mathbf{y}})$,
discrete text can be extracted from $\tilde{\mathbf{y}}$ using different heuristics~\citep{kumar2021controlled, qin-etal-2020-back,song-etal-2020-adversarial}. This formulation has been studied in various generation settings in prior work with different instantiations of $\mathbf{\tilde{y}}$ and $\mathcal{E}(\mathbf{y})$.

% with  either defined using a linear combination of the LM objective and the constraints with pre-define weights [CITE COLD, DELOREAN] or using a Lagrangian [CITE MUCOCO] which allows for adaptive weights for each constraint.  
However, this setup is deterministic and does not facilitate sampling.\footnote{While initialization can be used to add randomness to this algorithm, we find that it has little to no effect on diversity.} In addition, representing each token with a vector of size $|\mathcal{V}|$ can be computationally very expensive and difficult to fit into commonly used GPUs for long sequences (with more than ${\sim}$20-30 tokens; \Sref{subsec:discussion}). 

\section{Constrained Sampling via Langevin Dynamics in Embedding Space}
\label{sec:method}
To enable efficient gradient-based sampling from LMs,
% solve the aforementioned issues, 
we introduce \ourmodel which modifies the non-autoregressive framework in \Sref{sec:background} to (a) generate multiple samples instead of optimizing for only one deterministic output, (b) optimize for much smaller intermediate token representations as opposed to their distribution on the entire vocabulary. 
%First, we describe our proposed way to representing tokens followed by how they can facilitate sampling.
\subsection{Exploring the token representation space}
\label{subsec:pgd}
Instead of relaxing each target token $y_n$ as a soft representation over the vocabulary $\tilde{y}_n \in \mathbb{R}^{|\mathcal{V}|}$, we represent it as $\tilde{e}_{n} \in \mathbf{E}$. Here $\mathbf{E}$ denotes the embedding table of the underlying language model containing $|\mathcal{V}|$ vectors of size $d \ll |\mathcal{V}|$. We denote this sequence of embeddings as $\mathbf{\tilde{e}} = \{\tilde{e}_{1}, \ldots, \tilde{e}_{N}\}$.
At an update step $t$, instead of feeding each $\mathbf{\tilde{y}}$ to the model(s) (which are then transformed to an embedding to be fed to the first layer), we directly feed $\mathbf{\tilde{e}}$ to the first layer to compute the energy function, now defined as a function of embeddings instead of tokens.
% $\mathcal{E}(\mathbf{\tilde{e}_y})$
In case of deterministic minimization (similar to \eqref{eq:expgd}), these vectors are updated as
\begin{align}
    \mathbf{\tilde{e}}^{t} = \mathbf{\mathrm{Proj}}_\mathbf{E}(\mathbf{\tilde{e}}^{t-1} - \eta \nabla_{\mathbf{\tilde{e}}} \mathcal{E}(\mathbf{\tilde{e}}^{t-1})), 
    \label{eq:embedgd}
\end{align}
where $\mathbf{\mathrm{Proj}}_\mathbf{E}(\hat{e}) = \mathrm{arg}\min_{e \in \mathbf{E}} \|e - \hat{e}\|_2$ denotes a projection operation on the embedding table $\mathbf{E}$. In other words, after every gradient step, we project each updated vector back to a quantized space, that is the embedding table using Euclidean distance as the metric. This projection is done to prevent adversarial solutions.\footnote{Several prior works ~\citep{belinkov-glass-2019-analysis} have shown that neural-network based models are not robust to change in input space. We observed this phenomenon in our preliminary experiments where, without any projection, most low energy solutions were found to be garbled text.} 
After the optimization is complete, discrete text can be easily obtained by projection, that is the token indices corresponding to each $\tilde{e}_{n}$ in the embedding table $\mathbf{E}$. This formulation yields the following benefits: (a) For a sequence of length $L$, at any optimization step $t$, it only maintains (and computes gradients with respect to) $L\times d$ parameters, as opposed to $L\times |\mathcal{V}|$. This enables us to store much longer sequences in a GPU as compared to the storing $\mathbf{\tilde{y}}$. 
% In addition, we observe considerable improvements in speed for each gradient step\footnote{Because of the projection step in \eqref{eq:embedgd}, although the theoretical computational complexity of each gradient update is the same as \eqref{eq:expgd}, gradient computation graphs computed by libraries like P}, 
(b) this formulation provides a natural way to define hard rule-based constraints based on keywords or phrases (discussed in more detail in \Sref{sec:hard-constraints}), and, finally (c) it yields a natural way to generating samples.

\subsection{Gradient based Sampling via Langevin Dynamics}
\label{subsec:langevin}
The minimization in~\eqref{eq:embedgd} can be very easily extended to a sampling procedure by modifying the gradient descent in \eqref{eq:embedgd} to Langevin Dynamics~\citep{doi:10.1137/0329055,10.5555/3104482.3104568},
\begin{align*}
    \small
    \mathbf{\tilde{e}}^{t} = \mathbf{\mathrm{Proj}}_\mathbf{E}(\mathbf{\tilde{e}}^{t-1} - \eta \nabla_{\mathbf{\tilde{e}}} \mathcal{E}(\mathbf{\tilde{e}}^{t-1})  + \sqrt{2\eta\beta}z^{t})
    % \mathbf{e}_{y}^{t} = \mathbf{\mathrm{Proj}}_\mathbf{E}(\mathbf{e}_{y}^{t-1} - \eta \nabla_{\mathbf{e}_{y}} \mathcal{E}(\mathbf{y}))
    \label{eq:mucola}
\end{align*}
Langevin Dynamics provides a MCMC method to sample from a distribution using only the gradient of its logarithm. 
That is,
% if we wish to sample from $P(\mathbf{y} | \mathbf{x}; \theta)$ without constraints, we can use $\nabla_\mathbf{y} \log P$.
% In a constrained generation setting, 
if we define a distribution as $Q(\mathbf{y}) \propto \exp{(-\mathcal{E}(\mathbf{y}))}$, its logarithm leads to the update specified above.%in \eqref{eq:mucola}
\footnote{The normalization term in $Q(\mathbf{y})$ vanishes as its gradient with respect to $\mathbf{y}$ is $0$.} This method is often used for non-convex optimization for training neural networks~\citep{10.5555/3104482.3104568} due to its ability to escape local minima due to added noise and converge towards the global minima. In this work, we adapt it for inference~\citep{NEURIPS2019_3001ef25}.

Intuitively, by adding noise at every gradient step, this procedure intends to find outputs $\mathbf{y}$ that do not exactly minimize $\mathcal{E}$ but remain in the vicinity of the minima. In other words, it finds outputs which admit high probability under the distribution $Q(\mathbf{y})$. 
% Similar to other MCMC algorithms, 
This process begins with an exploration phase which is controlled by $\beta$. With a high value of $\beta$, the noise term is large leading to big updates. By gradual annealing such that $\beta \rightarrow 0$, as $t\rightarrow\infty$, this process converges to a sample from $Q(\mathbf{y})$.\footnote{
More details of the implementation of annealing schedule can be found in \Sref{sec:experiments}.
A similar noise can also be applied directly to the soft-token representations in \eqref{eq:expgd} as explored 
in~\citet{qin-2022-cold}. However, as we discuss in \Sref{subsec:discussion}, our formulation with its smaller parameter size allows generating longer sequences. In addition, considering logits as soft-representations (followed by softmax) has shown to result in slow mixing, that is, it takes much longer to converge as empirically shown in~\citet{hoang-etal-2017-towards} and also observed in~\citet{qin-2022-cold}. On the other hand, considering the simplex itself~\citep{kumar2021controlled, hoang-etal-2017-towards} as soft-representations is not compatible with Gaussian noise and can lead to undesirable behavior~\citep{DBLP:conf/nips/PattersonT13}.}

\subsection{Representing the energy function}
\label{subsec:lagrangian}

% Our goal in this work is sampling from the LM $P$ with a given set of controls. 
A straightforward way to represent $\mathcal{E}$ is with a linear combination as $\sum_{i=1}^C \lambda_i f_i - \lambda_{C+1} \log P$, with pre-defined weights $\lambda_1, \ldots, \lambda_{C+1}$
%This has been used in prior work in controlled text generation
~\citep{hoang-etal-2017-towards,qin-etal-2020-back,qin-2022-cold}. With this formulation
% This formulation has two issues: 
(a) linear weights, $\lambda_i$'s, can be hard to define and tune for different $f_i$, and especially difficult when $f_i$'s lie on different scales, and more importantly, (b) defining the energy function in this manner modifies the original goal, which is to sample from the language model $P$ (with constraints), not from a modified distribution $Q \propto \exp(-\mathcal{E} (\mathbf{y}))$. To alleviate these issues, we define the inference objective, following  \citet{kumar2021controlled}, as
\begin{align*}
    \mathbf{y} &\sim P(\mathbf{y}|\mathbf{x}; \theta)\text{, subject to } f_i([\mathbf{x}], \mathbf{y}) \leq \epsilon_i \forall i
     %f_i([\mathbf{x}], \mathbf{y}) &\leq \epsilon_i, i \in \{1, \ldots, C\},
 \end{align*}
where each threshold $\epsilon_i$ is a hyperparameter. As we discuss in more detail in \Sref{sec:experiments}, these thresholds can be flexibly defined for most kinds of constraints. For example, instead of merely trying to reduce $p_\textsc{toxic}(\mathbf{y})$,
% , depending on the underlying application, 
we can set it as $p_\textsc{toxic}(\mathbf{y}) < 0.1$. Given this formulation, we define the energy function as a Lagrangian 
% \begin{align}
    $\mathcal{E}(\mathbf{y}) = -\log P (\mathbf{y}) - \sum_{i=1}^u \lambda_i (\epsilon_i - f_i(\mathbf{y}))$.
    % \label{eq:energy}
% \end{align}
Here, $\lambda_i\geq0$ are Lagrangian multipliers and dynamically updated at each step. We follow the gradient of $\mathcal{E}$ downwards for the $\mathbf{\tilde{e}}$ (as described in \eqref{eq:embedgd}) and upwards for the multipliers (gradient ascent without any noise) while making sure that the multipliers remain positive:
% (by setting the multipliers to 0 whenever they become negative):
% \begin{align*}
    $\lambda_i^t = \max(0, \lambda_i^{t-1} + \alpha \nabla_{\lambda_i} \mathcal{E}(\mathbf{y}))$ (%\end{align*}
$\alpha>0$ is the step size for ascent). Intuitively, if a constraint is not satisfied, the term $(\epsilon_i - f_i(\cdot))$ would be negative and $\lambda_i$ would keep increasing making $\mathcal{E}$ high. 
On the other hand, if all the constraints are satisfied these values gradually decrease to $0$ making $\mathcal{E(\mathbf{y})} = -\log P(\mathbf{y})$ making the final output a sample from the desired distribution $P$. 
We implement a damped version of this process to improve stability, the details of which can be found in~\citet{kumar2021controlled}.
The final decoding algorithm we used in our experiments is described in algorithm \ref{algo} in the Appendix.

% [Should I discuss gumbel softmax here or somethere else?]
% 1. Describe regular decoding from language models, describe notations like word embeddings DONE

% 2. Convert that to sampling (talk about langevin dynamics, gumbel max). Compare and contrast with autoregressive sampling

% 3. Comment on speed

% 4. Proposed solution: word embedding optimization, and langevin dynamics in word embedding space. Describe loss functions.
\noindent
\textbf{Energy as a function of embeddings}
Performing gradient updates with respect to $\mathbf{\tilde{e}}$ requires that all objectives be defined as functions of $\mathbf{\tilde{e}}$, 
% as opposed to the output sequence 
not $\mathbf{y}$. 
Also, $f_1(\mathbf{y}), \ldots, f_C(\mathbf{y})$ must share the same input embedding table (as that of $P$).
We discuss in \Sref{sec:experiments} how this can achieved for different kinds of constraint functions $f_i$.
First, we describe how to compute the primary objective $-\log P(\mathbf{y}|\mathbf{x}; \theta)$ and its gradients with respect to $\mathbf{\tilde{e}}$.
In typical LMs, this objective
% is computed using a language model $P$ 
is factorized as $\log P(\mathbf{y}|\mathbf{x}) = \sum_{n=0}^{L-1} \log P(y_{n+1} | y_{1:n}, \mathbf{x})$. 
% where each $y_n$ is the one-hot vector of the $n$-th token in the sequence. 
For each decoding step $n+1$: the model takes as input $y_n$, which is converted to $e_n$ via an embedding table lookup. Passed through the network layers, it is converted to a hidden vector $h_n$. Since the input and output embedding tables in most modern LMs are shared~\citep{radford2019language,2020t5,lewis-etal-2020-bart,gpt3},\footnote{Even if the embedding tables are not shared, this loss may be computed and optimized using vectors from the output embedding table as parameters without any significant loss in performance.} the softmax probability is computed as,
% \begin{align}
    $P(y_{n+1} | y_{1:n}, \mathbf{x}) = \frac{\exp(h_n^Te_{n+1} + b_{n+1})}{\sum_{j=1}^{|\mathcal{V}|} \exp(h_n^Te_j + b_j)}$
    % \label{eq:ce}
% \end{align} 
where $b_n$ are optional bias terms. By replacing $e_{n+1}$ with $\tilde{e}_{n+1}$, we convert the above probability to $P(\tilde{e}_{n+1} | \tilde{e}_{1:n}, \mathbf{x})$. For each position $n+1$, $\tilde{e}_{n+1}$ receives gradients, (a) directly from $-\log P$ function %(it appears in both the numerator and the denominator),
and (b) through $h_{n+1}$ via back-propagation through the network layers (See \fref{fig:primary-objective} (left)).

% can be computed by replacing $e_{i+1}$ with $\tilde{e}_{i+1}$ in the above equation. 
\ignore{
\begin{figure*}
    \centering
    \includegraphics[width=\textwidth]{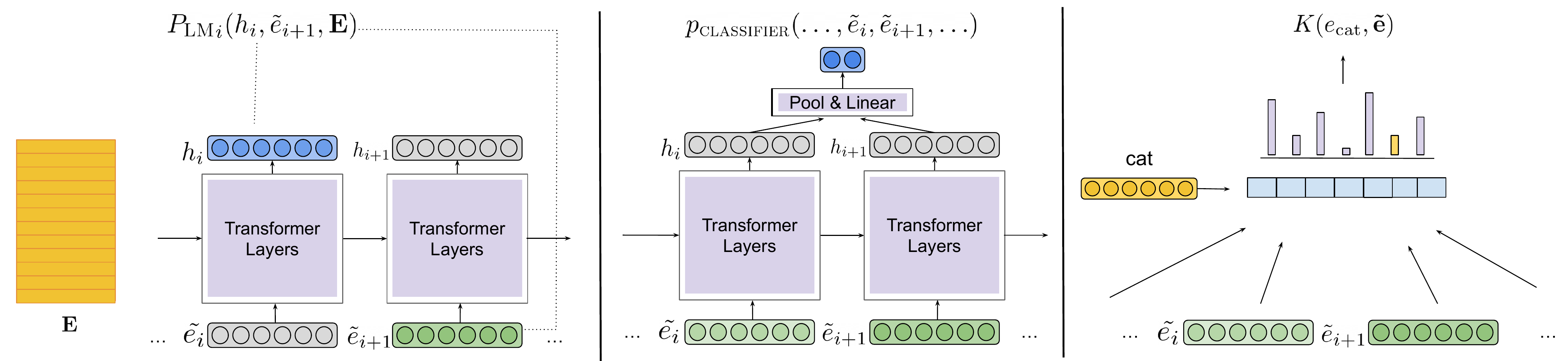}
    \caption{Different kinds of functions can be incorporated into \ourmodel defined on a shared embedding table $\mathbf{E}$. (Left) Language Modeling objective defines a per-token loss directly on the sequence of embeddings. For every token this loss provides gradients to update $\tilde{e}_i$ via backpropagation through the transformer layers and directly to $\tilde{e}_{i+1}$ through the negative loss likelihood loss as computed in \eqref{eq:ce}. This is used as a primary objective for the underlying LM and can also be used for classification as discussed in \Sref{subsec:sentiment-controlled-generation} (Center) Classification objective defined on probability of the desired label. The classifier gets the token embeddings $\tilde{\mathbf{e}}$ directly as input and updates the embedding using gradients obtained via backpropagation from the transformer layers (Right) Lexical loss defined on the embeddings directly (without the use of additional models) to include desired keywords or phrases in the output sequence (\Sref{sec:hard-constraints}). In practice any combination of these constraints can be used.}
    \label{fig:primary-objective}
\end{figure*}
}

\section{Experimental Setup}
\label{sec:experiments}
We evaluate \ourmodel on four constrained generation tasks. These tasks are selected based on defining different kinds of constraints for which prior work designed specialized training or decoding mechanisms which cannot be generalized beyond those tasks or language models. The main contribution of \ourmodel is generating diverse samples which conform to the language model $P$ as well as can satisfy user defined arbitrary combination of constraints for which fine-tuning is generally infeasible and tuning weights of each constraint is cumbersome. 
% \subsection{Implementation Details}
% \label{subsec:implementation}
For a pre-defined sentence length $L$, we initialize the token representation for each step $\tilde{e}_{1}, \ldots, \tilde{e}_{L}$ using token embeddings randomly sampled from the target vocabulary $\mathcal{V}$.\footnote{We also tried other initialization strategies like initializing with zeros, or outputs of nucleus sampling or greedy decoding but did not find it to have any significant effect on the final output} For all our experiments, we run the Langevin Dynamics simulation for a maximum of $250$ iterations unless specified otherwise. We describe additional implementation details including noise schedule, stopping criterion and multiplier update schedule in Appendix~\ref{appsec:implementation-details}.
% [Rest of this section is the biggest candidate for appendix]

\section{Text Generation with Soft Constraints}
\label{sec:soft-constraints}
First, we evaluate \ourmodel with real valued constraint functions defined via auxiliary models such as classifiers or LMs. Given an LM GPT2-Large~\citep{radford2019language}, and a prompt $\mathbf{x}$, we generate continuations $\mathbf{y}$. We conduct experiments with: toxicity avoidance and sentiment control. Each of the tasks define a binary constraint. Let the desired label be denoted by $\textsc{label}_1$, and other one with $\textsc{label}_0$ ($\textsc{label}_1$ is non-toxic in toxicity avoidance and positive in sentiment control). For both setups, we assume availability of corpora to train the constraint functions.\footnote{This setup can be easily extended to $n$-class setups by defining $n-1$ constraints as $p_0 > p_1, \ldots, p_0 > p_{n-1}$}
\paragraph{Baselines}
In addition to decoding without any constraints (which we simply call GPT2), we consider the following baselines which decode from left-to-right:%, and define constraint functions derived from them.

\begin{compactitem}
\item \textbf{Domain Adaptive Pretraining (DAPT)} \citep{gururangan-etal-2020-dont} proposes to finetune the LM $P$ on a corpus of desired constraint and sample directly from finetuned version.

\item \textbf{FUDGE} \citep{yang-klein-2021-fudge} uses a ``future-aware'' constraint classifier to modify output token probabilities at every decoding step to steer the generation to promote constraint satisfaction. This classifier is trained to predict the ground truth label for every prefix of the training corpus. %We train this classifier on the same dataset we use to train our constraint classifier. And 
% We use the recommended hyperparameters for decoding with top-$k$ sampling with $k=10$.

\item \textbf{GeDi}~\citep{KrauseGeDi2020} uses a class-conditioned LM to modify output token probabilities at each step via Bayes’ rule. %We use two versions of this baseline with class conditional LMs trained on the described datasets (SST-2 and Yelp).

\item \textbf{DExperts} \citep{liu-etal-2021-dexperts} proposes to replace the class-conditioned LM with two auxiliary language models (one expert and one anti-expert) to modify the output logits at every step. These LMs are trained using same setup as the baseline \textbf{DAPT} but instead of sampling from them directly, it uses them to steer the base LMs outputs.
%we use the fine-tuned LMs used in \textsc{DExperts} directly to generate positive or negative continuations. 
% \paragraph{FUDGE~\citep{yang2021fudge}}: We train classifiers on SST-2 and Yelp dataset in a similar fashion as described in \Sref{subsec:toxicity} and decode with recommended generation hyperparameters.
For each of the baselines, we use recommended hyperparameters to generate samples.

\end{compactitem}

\paragraph{Constraint functions}

Each of these baselines can be adopted as constraint functions for \ourmodel as follows:

\begin{compactitem}
\item \textbf{Discriminative Classifiers} 
% Similar to toxicity avoidance (\Sref{subsec:toxicity}), 
We train a binary classifier $p_\textsc{label}(\mathbf{y})$, which predicts the probability of the desired attribute given the output sequence $\mathbf{y}$ by finetuning  \href{https://huggingface.co/roberta-base}{roberta-base} with GPT2-Large embeddings (more details in \Aref{appsec:training-details}). To decode with \ourmodel, we formulate this constraint as $p_{\textsc{label}_1} > \epsilon$ (We define specific $\epsilon$ values in \Sref{subsec:toxicity} and \Sref{subsec:sentiment-controlled-generation} respectively). To improve its gradient profile, we use the constraint in log space.
% $\log p_{\textsc{label}_1} > \log \epsilon$.
We call this setup \textbf{\ourmodel-\textsc{disc}}.

% positive sentiment, $p_\textsc{sentiment}(\cdot)$ on the full sequence (including the prompt). % We train two versions of this classifier on two datasets: SST-2 corpus~\citep{socher-etal-2013-recursive} containing ${\sim}4$K examples in Movie domain for each class; and Yelp polarity corpus containing ${\sim}280$K examples for each class containing a mixed domain of reviews. 
% Again, we train both of these classifiers by finetuning \href{https://huggingface.co/roberta-base}{roberta-base} with GPT2-Large embeddings (more details in the Appendix \Aref{appsec:training-details}). We use the constraint $p_\textsc{positive}([\mathbf{x}, \mathbf{y}]) < 0.1$ or $p_\textsc{positive}([\mathbf{x}, \mathbf{y}]) > 0.9$ depending on the desired polarity (in practice we use their logarithm). We refer to it as \ourmodel-\textsc{disc}. We also test a setup which uses both of the classifiers as two separate constraints \ourmodel-\textsc{two-disc}.

\item \textbf{Generative Classifiers}
Prior work has shown that discriminative classifiers can be fragile to domain shift or adversarial examples~\citep{https://doi.org/10.48550/arxiv.1703.01898,krause2020gedi}. Hence, we also consider a second class of \emph{generative} classifiers trained as class conditional LMs that model $p(\cdot| \textsc{label})$. % where \textsc{sentiment} can be either positive or negative.  
Intuitively, they are required to explain every word in the input, potentially amplifying the class signal and improving robustness~\citep{min2021noisy}. We define them in three ways by finetuning GPT2-Large: (1) following \textsc{GeDi} (\textbf{\ourmodel-GeDi}), %keeping the embedding table frozen (more details in \Aref{appsec:training-details}). 
% we concatenate the word "positive" in front of every positive example and "negative" in front of every negative example and train an LM by finetuning GPT2-Large (more details in the Appendix \Aref{appsec:training-details}). 
% We call this setup \ourmodel-\textsc{gen}. 
(2) following DExperts, %~\citep{liu-etal-2021-dexperts},
(we train two separate LMs; \textbf{\ourmodel-DExperts}). 
% one for positive class and one for negative class. 
% We call this setup \ourmodel-Dexperts. 
%For both cases, we set the constraint as $p([\mathbf{x}, \mathbf{y}] | \textsc{sentiment}=\textsc{positive}) > p([\mathbf{x}, \mathbf{y}] | \textsc{sentiment}=\textsc{negative})$ for positive sentiment and vice-versa for negative (again, we realize the constraints in log-space for better gradients). 
% \paragraph{Prompt based Classifier}
And finally, (3) %we also consider a ``classifier'' without any finetuning.
motivated by recent work on prompt-based classification, we define a class-conditional LM without finetuning the model as $P(\mathbf{x}, \mathbf{y} | \text{verbalize}(\textsc{label}))$ where $\text{verbalize}(\cdot)$ is function that converts the label to a natural language string (\textbf{\ourmodel-prompt}). Note that for all three setups, the embedding table is frozen (more details in \Aref{appsec:training-details}). We decode via \ourmodel with the constraint 
% we consider a very simple constraint using the underlying LM itself:
$p(\mathbf{x}, \mathbf{y} | \textsc{label}_1)) > p(\mathbf{x}, \mathbf{y}| \textsc{label}_0)$ (again, realized in log-space)\footnote{Note that all constraints we descibe can be easily extended to $n$-class set (with say $0$ as the desired label)  by defining $n-1$ constraints as $p_0 > p_1, \ldots, p_0 > p_{n-1}$}.
% where $\text{verbalize}()$ is function that converts the label to a natural language string. That is, we simply append the sequence ``This is amazing (terrible)'' in front of the output sequence and predict the probability of the rest of the sequence. 
% We refer to this as \ourmodel-\textsc{prompt}. We use the same constraint format here as in generative classifiers. 
\end{compactitem}

\paragraph{Evaluation}
% \paragraph{DExperts~\citep{liu-etal-2021-dexperts}}: We use two language models trained on the positive and negative corpus (for both SST-2 and Yelp) and decode with recommended hyperparameters.
In both experiments, we evaluate the generated samples along three dimension, (1) \textbf{Constraint Satisfaction} measured using external evaluators, (2) \textbf{Fluency}, measured by mean perplexity of the continuations measured using GPT2-XL. Since the objective is to generate samples from the LM, we rank different methods not by their absolute perplexity, but its difference from the perplexity of unconstrained text. Additionally, we also report a grammaticality score: the fraction of outputs predicted by a classifier trained on CoLA~\citep{warstadt-etal-2019-neural} as fluent. (3) \textbf{Diversity}, measured by computing the mean number of distinct n-grams in each set of samples, normalized by the length of text~\citep{li-etal-2016-diversity}. We report this for $n=1,2,3$ following prior work. Since all the automatic metrics are model based and can be biased, we also perform human evaluation in an A/B testing setup with the best performing baseline (DExperts in our case). For each sample, we ask 3 annotators to compare and rank candidates from our approach and the baseline on constraint satisfaction, topicality and fluency. 

\subsection{Toxicity Avoidance}
\label{subsec:toxicity}
Prior work have shown that large pre-trained LMs are at risk of producing toxic content even when given innocuous prompts~\citep{sheng-etal-2019-woman, gehman-etal-2020-realtoxicityprompts}. In this experiment,
% we apply \ourmodel to steer the LM outputs away from such behavior. 
given a neutral prompt, we generate non-toxic continuations using \ourmodel. 
% This task has been widely studied with solutions ranging from finetuning with non-toxic data~\citep{}, blocklisting ``toxic'' keywords~\citep{}, or using classifiers~\citep{} or additional language models t~\citep{}. 
We only consider the setup \ourmodel-\textsc{disc} here, with a classifier $p_\textsc{toxic}$, trained on a dataset of human-annotated comments labelled as toxic or non-toxic (\Aref{appsec:training-details}). We decode with the constraint $p_\textsc{toxic} < 0.01$.

\ignore{We use GPT2-Large~\citep{radford2019language} as our underlying LM and define a single constraint based on a binary classifier $p_{\textsc{toxic}}$ to measure toxicity. 
This classifier is trained on a dataset of human-annotated comments from the \href{https://bit.ly/3cvG5py}{Jigsaw Unintended Bias in Toxicity Classification Kaggle challenge}. The dataset has ${\sim}160$K toxic comments and ${\sim}1.4$M nontoxic comments. This task setup was introduced in~\citet{liu-etal-2021-dexperts} and details of how this dataset is created can be found in the paper. We train this classifier by finetuning \href{https://huggingface.co/roberta-base}{roberta-base}~\citep{liu2019roberta} by simply replacing its embedding table with the one from GPT2-Large. Please refer to \Aref{appsec:training-details} for more details. This classifier predicts the toxicity probability of an output sequence $\mathbf{y}$. We use the constraint $p_\textsc{toxic}(\mathbf{y}) \leq 0.01$ in this experiment. To improve its gradient profile, we modify it as $\log (p_\textsc{toxic}) \leq \log(0.01)$. }
% \subsubsection{Baselines}
We follow the evaluation setup defined in~\citet{liu-etal-2021-dexperts} and use test set of 10K nontoxic prompts% from the RealToxicityPrompts dataset
~\citep{gehman-etal-2020-realtoxicityprompts} where without any constraints, the user might receive harmful output from the LM. For each prompt, we generate $25$ samples for length $20$ tokens each. We measure constraint satisfaction using the toxicity score from Perspective API. Following prior work~\citep{gehman-etal-2020-realtoxicityprompts, liu-etal-2021-dexperts}, we report the maximum toxicity score over 25 samples per prompt averaged over the number of prompts, and the empirical probability of generating a continuation with toxicity $>0.5$ at least once over the 25 generations.%We consider two main baselines: 

\begin{table*}[]
\small
\centering
\begin{tabular}{@{}lccccccc@{}}
\toprule
\multirow{2}{*}{\textbf{Approach}} & \multicolumn{2}{c}{\textbf{Toxicity}} & \multicolumn{2}{c}{\textbf{Fluency}}& \multicolumn{3}{c}{\textbf{Diversity}} \\ \cmidrule(l){2-8} 
& \multicolumn{1}{c}{\begin{tabular}[c]{@{}c@{}}Avg. Max.\\Toxicity\end{tabular}} & \multicolumn{1}{c}{\begin{tabular}[c]{@{}c@{}}Toxicity\\ Prob.\end{tabular}} & \multicolumn{1}{c}{Perplexity} & \multicolumn{1}{c}{\begin{tabular}[c]{@{}c@{}}CoLa\\Accuracy\end{tabular}} & \multicolumn{1}{c}{Dist-1} & \multicolumn{1}{c}{Dist-2} & \multicolumn{1}{c}{Dist-3} \\ \midrule
\textsc{GPT-2} & 0.527 & 0.520  & 25.45 & 88.3  & 0.58   & 0.85   & 0.85 \\ \midrule
%\textsc{PPLM}  & 0.520 & 0.518 & 32.58 &  86.7  & 0.58   & 0.86   & 0.86 \\
\textsc{DAPT}  & 0.428 & 0.360  & 31.21 &  \textbf{91.2}  & 0.57   & 0.84   & 0.84 \\
\textsc{FUDGE} & 0.437 & 0.371 & 12.97 & 88.5 & 0.47 & 0.78 & 0.82 \\
\textsc{GeDi}  & 0.363 & 0.217 & 60.03 & 85.5    & 0.62   & 0.84   & 0.83 \\
% DEXPERTS (anti-only) & 0.352 & 0.191 & 52.02 &    & 0.58   & 0.8    & 0.73 \\
\textsc{DExperts} & \textbf{0.302}   & 0.118 & 38.20  & 89.8 & 0.56 & 0.82 & 0.83 \\ \midrule
\ourmodel  & \textbf{0.308} & \textbf{0.088}  & \textbf{29.92} & 88.2 & 0.55 & 0.82 & 0.83 \\ \bottomrule
\end{tabular}
\caption{Results for toxicity avoidance (\Sref{subsec:toxicity}). We evaluate on three axes: (1) Toxicity--Avg. Max. Toxicity and Toxicity Prob.: lower the better. (2) Fluency--GPT2-XL Perplexity: the closer the value to unconstrained outputs (GPT2: 38.6), the better; CoLa accuracy: higher the better, and (3) Diversity (Dist-1,2,3): higher the better. The best values in each column are highlighted in \textbf{bold}. While our method improves or performs on par with baselines on toxicity metrics, we obtain substantial improvements on perplexity.}
\label{tab:toxicity-avoidance-results}
\end{table*}

As shown in \Tref{tab:toxicity-avoidance-results}, \ourmodel outperforms or matches all baselines on toxicity, including a strong baseline \textsc{DExperts} which is specifically designed for binary constraints. 
%on toxicity probability. 
In addition, our method is closest in perplexity to unconstrained generation, while maintaining grammaticality as well as diversity of baseline methods\footnote{While FUDGE obtains the lowest absolute perplexity, prior work~\citep{Holtzman2020The} has shown that very low perplexity is not an indicator of higher quality but of repetitions and usage of only high frequency tokens.}.  We attribute this improvement to the fact that after the constraints are satisfied, the energy function in \ourmodel reduces to $-\log P(\mathbf{y})$, the original function we intend to sample from, whereas in the baselines, the underlying probability distribution (or the energy function) is modified to achieve control. Furthermore, human evaluation (\Aref{appsec:additional-results}) reveals that generations by \ourmodel match DExperts on toxicity and fluency while being more topical.

% For human evaluation, we sample a subset of 200 prompts from the test set with two generations per prompt and compare \ourmodel and DExperts (a total of 400 outputs per approach). We find [write human eval results]. 

% \subsubsection{Human Evaluation}
% TODO

%#######
\subsection{Sentiment Controlled Generation}
\label{subsec:sentiment-controlled-generation}

Given a prompt $\mathbf{x}$, the goal of this task is to generate continuations $\mathbf{y}$ using an LM with a positive sentiment/polarity.
% desired sentiment/polarity (either positive or negative). 
% With GPT2-Large as the underlying LM, we consider different ways to represent the sentiment constraint considered in prior work on controlled text generation. 
To understand the effect of sources of training data, we train two versions of each constraint function described above on two datasets: SST-2 corpus~\citep{socher-etal-2013-recursive} containing ${\sim}4$K examples in Movie reviews for each class; and Yelp polarity corpus containing ${\sim}280$K examples for each class containing a mixed domain of reviews. We also consider an additional setup where we use two constraints using both versions of \textbf{\ourmodel-\textsc{disc}}, which we call \textbf{\ourmodel-\textsc{two-disc}}.
We use a dataset of 15 prompts from~\citet{Dathathri2020Plug} and generate $20$ samples per prompt of length $12$, $20$, and $50$.% for both positive and negative polarity. %We consider the same baselines as in \Sref{subsec:toxicity}.

\ignore{
\paragraph{Domain Adaptive Pretraining (DAPT)~\citep{gururangan-etal-2020-dont}} Here we use the fine-tuned LMs used in \textsc{DExperts} directly to generate positive or negative continuations. 

\paragraph{FUDGE~\citep{yang2021fudge}}: We train classifiers on SST-2 and Yelp dataset in a similar fashion as described in \Sref{subsec:toxicity} and decode with recommended generation hyperparameters.

\paragraph{GeDi~\citep{KrauseGeDi2020}} uses a class-conditioned LM to modify output token probabilities via Bayes’ rule. We use two versions of this baseline with class conditional LMs trained on the described datasets (SST-2 and Yelp).

\paragraph{DExperts~\citep{liu-etal-2021-dexperts}}: We use two language models trained on the positive and negative corpus (for both SST-2 and Yelp) and decode with recommended hyperparameters.
}
% [could be removed? or moved to appendix?]
To evaluate constraint satisfaction, we measure positive sentiment accuracy of the output using three external classifiers to account for domain differences in their training data% Sentiment classifiers are trained on human-written texts in specific domains, which can make them fragile while evaluating machine generated text, and prone to be fooled by adversarial solutions~\citep{song-etal-2020-adversarial}. 
% Hence, we report accuracies measured three different classifiers trained on different datasets
% , two of them used in prior work
, (a) \textbf{\textsc{c1}}: \href{https://huggingface.co/distilbert-base-uncased-finetuned-sst-2-english}{distilbert}~\citep{Sanh2019DistilBERTAD} finetuned on SST-2 data, used in \citep{liu-etal-2021-dexperts}, (b) \textbf{\textsc{c2}}: \href{https://huggingface.co/textattack/bert-base-uncased-yelp-polarity}{bert-base}~\citep{devlin-etal-2019-bert} finetuned on Yelp Polarity corpus used in~\citet{mix-and-match-2022}, and (c) \textbf{\textsc{c3}}: \href{https://huggingface.co/siebert/sentiment-roberta-large-english}{SieBERT}~\citep{heitmann2020} finetuned on 15 different polarity datasets. 
We report a subset of results of this experiment in \tref{tab:sentiment-results-20-main} for outputs of length 20 (and the rest in \Aref{appsec:additional-results}). We observe a significant variance in sentiment control accuracies (\textsc{c1}, \textsc{c2} and \textsc{c3}) where constraints trained on SST-2 perform worse on the evaluator trained on Yelp (\textsc{c2}) and vice versa for all methods. The third evaluator (\textsc{c3}) trained on a much larger training set can be considered more reliable. Overall, we find that \ourmodel in all settings obtains perplexity values closer to unconstrained outputs (GPT2) whereas most baselines achieve control at the cost of perplexity. % This behavior is expected since the baselines modify the underlying LM distribution while introducing constraints. 
 Surprisingly, constraints trained on Yelp perform poorly compared to those trained on SST2 despite the former being a larger dataset.

For outputs of lengths $12$ and $20$, 
% for positive control, 
% we find that 
%\ourmodel using both discriminative classifiers together 
\ourmodel-\textsc{two-disc}
finds a good balance of control and fluency and  outperforms all other baselines on sentiment accuracy while maintaining good perplexity (except \textsc{GeDi} which performs poorly on perplexity as well as CoLa accuracy). This improvement however comes with a slight decline in diversity metrics which we argue is a fair price to pay for constraint satisfaction compared to fluency. Similar to \Sref{subsec:toxicity}, a small scale study on human evaluation (\Aref{appsec:additional-results}) reveals \ourmodel to be more topical than the best baseline DExperts. Finally, using a prompt-based constraint also performs strongly despite not trained at all. In future work, we will look into training a prompt-based classifier to improve this performance. 
% For negative control, \ourmodel has a similar trend and beats all baselines except \textsc{DExperts}. This is despite the fact the for all our setups, we observed a constraint satisfaction rate is $>95\%$ while decoding. This behavior warrants are deeper look into the robustness of the constraint models and we leave this for future work.
%It is noteworthy, however, that \textsc{DExperts} can only be used in prompt-based models and not conditional generation models \footnote{rephrase this nicely}. 
% However, f
For outputs of length $50$, we observe a slight drop in \ourmodel's performance. On closer inspection (\tref{tab:sentiment-examples50}), we find a trend of degenerate repetitions at the end of many sequences. Prior work~\citep{Holtzman2020The} has shown that large LMs often assign unusually high probabilities to repeating sequences especially with increasing lengths and since our method is designed to sample high probability outputs, such behavior is expected. In future work, we will explore constraints designed to discourage this behavior~\citep{Welleck2020Neural,https://doi.org/10.48550/arxiv.2202.00666}. 

% We additionally perform human evaluations similar to \Sref{subsec:toxicity} on outputs of length 20 where, for each prompt in the test set, we sample 6 generations and compare \ourmodel and DExperts. We find [write human eval results] %(a total of 30 outputs per approach). We find [write human eval results].

% (1) There is considerable variance in sentiment classifier accuracies. This is because domains of the constraints don't match the domains of the evaluators. Using two different classifiers together helps. Using no separately trained constraints (just prompt based classifier) is able to perform considerably well given it's heuristically designed. 
% (2) \ourmodel is good at positive control but lags behind dexperts on negative control even though constraint satisfaction is really high. 
% (3) Overall, our models are much better at sticking to perplexity values closer to unconstrained generation (GPT2) whereas most baselines achieve control but have much higher perplexity. This is because they modify the underlying LM distribution. 
% (4) Our best performing method(s) do lead to a bit decrease in diversity (not sure how to justify). 
% (5) Our model still struggles with longer lengths where we still are not able to beat some baselines. 
% (2) Domains of the trained constrained models matter, using a constraint

% \paragraph{Human Evaluation}
% TODO

\section{Decoding with Hard Constraints}
\label{sec:hard-constraints}

In the previous two tasks, we explored how \ourmodel can be applied on soft constraints, defined via real valued functions like probabilities of classifiers or language models. Now, we consider a ruled-based constraint that a specific word or phrase \emph{must} appear in the generated text. Existing autoregressive solutions to this task have explored various strategies either based on explicitly modifying probabilities to up-weight desired words~\citep{pascual-etal-2021-plug-play}, or search-based strategies based on beam-search~\citep{lu-etal-2021-neurologic}. 
% In this work, inspired by~\citep{editinvariant2022}, 
In this work, we define a differentiable distance function $d(w, \mathbf{\tilde{e}})$ which measures overlap between desired word ($w$) and the output token embeddings $\mathbf{\tilde{e}}$ (we use the notation $w$ to refer to as the word itself and its index in the vocabulary interchangeably). We then propose a simple criterion to define a threshold $\epsilon$ that guarantees that if $d(w, \mathbf{\tilde{e}}) < \epsilon$, then $w$'s embedding appears in $\mathbf{\tilde{e}}$ (and by extension $w$ appears in $\mathbf{y}$). Inspired from~\citet{editinvariant2022,qin-2022-cold}, this distance is computed in three steps (1) define a ``probability distribution'' for each output token,
$\pi_n = \mathrm{softmax}(-\|\tilde{e}_n - e_1\|^2_2, \ldots, -\|\tilde{e}_n - e_{|\mathcal{V}|}\|^2_2)$ where $\{e_1, \ldots, e_{|\mathcal{V}|}\} \in \mathbf{E}$, (2) Define $q= \mathrm{gumbel}-\mathrm{softmax}(g_1/\tau, \ldots, g_N/\tau)$, where $g_n = \log \pi_{n, w}$, gumbel softmax provides a way to differentiably sample~\citep{JanGuPoo17} and $\tau$ is a hyperparameter, and finally, (3) $d(w, \mathbf{\tilde{e}}) = \sum_{n=1}^N - q_n g_n$. Intuitively, this function minimizes the Euclidean distance between one of the output embeddings (chosen with stochasticity) and $w$'s embedding, $e_w$. %[should I provide more intuition here? this is bit dense]
% \begin{align*}
%     g_n &= \log \pi_{n, w} \\
%     q &= \mathrm{gumbel{-}softmax}(g_1/\tau, \ldots, g_N/\tau) \\
%     d(w, \mathbf{\tilde{e}}) &= \sum_{n=1}^N - q_n g_n
% \end{align*}
% where and $\tau$ is a hyperparameter. 
% We use hard sampling here to ensure $q$ is one-hot. 
This function can be easily extended for phrases, $w = (w_1, \ldots, w_l)$ by defining $g_n = \frac{1}{l}\sum_{u=1}^l \log \pi_{w_u, n+u}$.
% $l$ distances and averaging them to compute each $g_n$, as $g_n = \sum_{u=1}^l \|e_{w_u} - \tilde{e}_{n+u}\|$. In other words, we consider every $l$ length subsequence in the output $\mathbf{\tilde{e}}$ and compute $g_n$ as the average of its Euclidean distance from corresponding tokens in the phrase $w$. 
% This computation can be efficiently done on a GPU using a convolution operation.

Based on this definition, for each desired keyword, we define a threshold $\epsilon_w$ as $-\log \pi_{w, w}$.% which ensures hard satisfaction. 
% That is, $w$ appears in the output text if and only if the constraint is satisfied. 
We provide an intuitive explanation of the distance function, and a semi-formal and empirical proof of hard satisfaction of this threshold in \Aref{appsec:keyword-constraint}.

\begin{table*}[]
\centering
\footnotesize
\begin{tabular}{@{}llcccccccc@{}}
\toprule
\multirow{2}{*}{\textbf{Approach}} &
\multirow{2}{*}{\textbf{Setting}} &
\multicolumn{3}{c}{\textbf{\% Positive Sentiment}} & \multicolumn{2}{c}{\textbf{Fluency}}& \multicolumn{3}{c}{\textbf{Diversity}} \\ \cmidrule(l){3-10} 
& & \multicolumn{1}{c}{\textsc{c1}}& \multicolumn{1}{c}{\textsc{c2}}& \multicolumn{1}{c}{\textsc{c3}}& \multicolumn{1}{c}{Perplexity} & \multicolumn{1}{c}{CoLa} & \multicolumn{1}{c}{Dist-1} & \multicolumn{1}{c}{Dist-2} & \multicolumn{1}{c}{Dist-3} \\ \midrule
\textsc{GPT-2} & - & 46.7 & 47.7 & 61.3 & 38.6 & 78.7 & 0.64 & 0.90 & 0.88 \\ \midrule
% \multirow{8}{*}{+}
\textsc{DAPT}  & SST-2 & 73.6 & 70.0 & 78.3 & 76.9 & 70.7 & 0.64 & 0.89 & 0.86 \\
\textsc{FUDGE} & SST-2 & 67.6 & 63.0 & 79.3 & 10.3 & \textbf{94.0} & 0.51 & 0.80 & 0.84 \\
\textsc{GeDi}  & SST-2 & 99.0 & 96.3 & 99.7 &  268.7 & 54.0 & 0.69 & 0.87 & 0.84 \\
\textsc{DExperts} & SST-2 & 91.2 & 83.4 & 95.4 & 55.37 & 81.6 & 0.61 & 0.89 & 0.87 \\ \midrule%\cmidrule(l){1-10}
\ourmodel-\textsc{disc} & SST-2 & 84.6 & 77.5 & 88.0 & 27.9 & 80.8 & 0.50 & 0.81 & 0.82 \\
\ourmodel-\textsc{disc} & Yelp & 83.0 & 83.6 & 83.0 & \textbf{32.2} & 76.0 & 0.50 & 0.75 & 0.80 \\ 
\ourmodel-\textsc{two-disc} & Yelp, SST-2 & \textbf{93.7} & \textbf{91.0} & \textbf{96.0} & 28.9 & 76.7 & 0.53 & 0.77 & 0.74 \\
\ourmodel-\textsc{prompt}  & - & 87.3 & 91.0 & 93.0 & 53.0 & 77.2 & 0.54 & 0.82 & 0.80  \\
% \midrule \\ %\midrule
% \multirow{8}{*}{-} & \ourmodel-\textsc{prompt}  & - & 12.7 & 11.3 & 14.7 & 30.3 & 77.7 & 0.48 & 0.76 & 0.80  \\
% & \ourmodel-\textsc{two-disc} & Yelp, SST-2 & 11.3 & 14.3 & 14.7 & 31.7 & 70.3 & 0.48 & 0.76 & 0.81 \\
% & \ourmodel-\textsc{disc} & Yelp & 17.7 & 23.0 & 30.7 & \textbf{32.4} & 70.0 & 0.55 & 0.80 & 0.84 \\ 
% & \ourmodel-\textsc{disc} & SST-2 & 16.0 & 21.0 & 25.7 & 45.6 & 69.3 & 0.56 & 0.83 & 0.85 \\ \cmidrule(l){2-11}
% & \textsc{DExperts} & SST-2 & \textbf{2.10} & \textbf{9.10} & \textbf{3.50} & 49.3 & 76.8 & 0.64 & 0.89 & 0.87 \\
% & \textsc{GeDi}  & SST-2 & 0.00 & 3.30 & 0.00 & 112.3 & 59.7 & 0.71 & 0.85 & 0.80 \\
% & \textsc{FUDGE} & SST-2 & 44.0 & 40.3 & 59.7 & 10.5 & \textbf{89.7} & 0.52 & 0.80 & 0.85 \\
% & \textsc{DAPT}  & SST-2 & 24.0 & 30.0 & 27.7 & 80.6 & 70.0 & 0.65 & 0.89 & 0.87 \\
\bottomrule
\end{tabular}
\caption{
Results for Sentiment Controlled Generation for outputs of length 20. We evaluate on three axes: (1) \% Positive Sentiment: higher the better. We use three external classifiers for this evaluation, \textsc{c1} trained on SST2 data, \textsc{c2} trained on Yelp data, and \textsc{c3} trained on 15 polarity datasets; (2) Fluency--GPT2-XL perplexity, closer the value to unconstrained outputs (GPT2: 38.6), the better; CoLa accuracy: higher the better, and (3) Diversity (Dist-1,2,3): higher the better. The best values in each column are highlighted in \textbf{bold}.}% [maybe move negative results to appendix?]
\label{tab:sentiment-results-20-main}
\end{table*}
\noindent
\textbf{Tasks }We formally evaluate this setup on two tasks: (1) open-ended keyword constrained generation (with two datasets: \textsc{CommonGen} and \textsc{ROC})), and (2) terminology constrained machine translation. We additionally show preliminary findings on a third task, entity guided summarization. We elaborate on \textsc{CommonGen} here and the rest of the results, following a similar trend can be found in \Aref{appsec:keyword-constraint}. In \textsc{CommonGen}~\citep{lin-etal-2020-commongen} given no prompt 
% (just the start of sequence token)
, the task is generate an output of maximum length $40$ which contains a given set of four or five words. We use GPT2-XL as the underlying LM in this setup with \textsc{Cold}~\citep{qin-2022-cold} as our main baseline. 
% and \citep{pascual-etal-2021-plug-play} respectively and their corresponding baselines and underlying language models (GPT2-XL and GPT2-Large). %For every prompt, we run \ourmodel 3 times and select the one which satisfies all the constraints and has the highest log-likelihood according to the underlying language models. 

% For both datasets, we report results on GPT2-Large and GPT2-XL.

\noindent
\textbf{Evaluation }
Following prior work, we measure the performance on two axes, (1) \textbf{Coverage}, measured by (a) count average number of keywords appearing in the output; and (b) percent, measuring the fraction of outputs which contain all the desired keywords. (2) \textbf{Fluency}, as measured by GPT2-XL perplexity and human evaluation, where on a sample of 200 outputs, we ask 3 annotators to rate each output on a 3-point likert scale. 
% 
% We report a measure of the coverage of
% the constraint words in generation as well as language fluency by evaluating the perplexity of the text (measured by GPT2-XL) and grammatically measured by \textsc{CoLa} classifier same as used in \Sref{subsec:toxicity}.
As reported in \tref{tab:commongen}, we outperform the best baseline on coverage. % for both \textsc{CommonGen}.% and \textsc{ROC}. 
% For \textsc{ROC} (\tref{tab:roc-results}), while we beat most baselines, we under perform the best baseline $K2T$ on coverage. We attribute this decline to that fact that we use our method only with a maximum length of $40$ as opposed to the maximum length $90$. This is again due to degenerate repetitive behavior observed when running our algorithm with a longer maximum length. With constraints designed to discourage repetitions this can potentially be alleviated. However,
 We outperform all baselines in terms of perplexity by a large margin, again owing to the fact that our method samples from the language model and does not modify the distribution itself as opposed to the baselines. Human evaluation reveals that our approach slightly underperforms the best baseline. 
% [need some explanation here]. 
%We perform additional experiments with keyword-constrained translation and a preliminary exploration of entity constraint summarization where we achieve similar performance improvements, the details of which can be found in \Aref{appsec:keyword-constraint}. 

% [remaining human evaluation]

%We also ask crowdworkers to rate the text fluency on a 3-point Likert scale on 200 test examples. The average ordinal Krippendorff alpha is 0.29, indicating a fair inner-annotator agreement.

% \usepackage{booktabs}
% \usepackage{multirow}
\begin{table}[]
\small
\centering
\begin{tabular}{@{}lrrrr@{}}
\toprule
\multirow{2}{*}{}        & \multicolumn{2}{c}{\textbf{Coverage}}                            & \multicolumn{2}{c}{\textbf{Fluency}}        \\ \cmidrule(l){2-5} 
                         & \multicolumn{1}{c}{Count} & \multicolumn{1}{c}{Percent} & \multicolumn{1}{c}{Perplexity} & \multicolumn{1}{c}{Human}  \\ \midrule
TSMH                     &   2.72                        & 71.27                       & 1545.15   & 1.72                      \\
Neurologic               & 3.30                         & 91.00                          & 28.61 & \textbf{2.53}                      \\
COLD                     & 4.24                 & 94.5               & 54.98 &  2.07                 \\
% \ourmodel & 4.07                 & 93.8               & 31.48 & 2.35 \\
\ourmodel & \textbf{4.49}                & \textbf{99.7}               & \textbf{23.50} & 2.29  \\ \bottomrule
\end{tabular}
\caption{Results of keyword constraint on \textsc{CommonGen}. We report (a) coverage as avg. count of desired keywords in the output and the fraction of the outputs containing all keywords (percent); and (b) GPT2-XL perplexity and avg. fluency score rated by humans.}
\label{tab:commongen}
\end{table}

\section{Discussion and Analysis}
\label{subsec:discussion}

% \noindent
% \textbf{Speed and Memory}
\paragraph{Speed and Memory}
% \paragraph{Speed and Memory Requirements} 
% Before discussing large scale experiments using our proposed method, we first describe a toy-setting to study the impact of using token embeddings as optimizable parameters instead of vocabulary sized token representations.
% As we discussed in \Sref{subsec:pgd},
Generating a sequence of length $L$ using \ourmodel requires maintaining $L\times d$ parameters. In contrast, performing Langevin Dynamics in the vocabulary space requires $L\times |\mathcal{V}|$ parameters ($|\mathcal{V}| >> d$). In this analysis, we empirically verify the benefits of our setup. With GPT2-Large as underlying LM, we sample sequences of varying lengths with various constraints, on different commercially available GPUs using both our approach and an ablation with vocabulary sized representations (logits+softmax; more details in \Aref{appsec:additional-discussion}). 
As summarized in \tref{tab:speed-analysis}, we find that much longer sequences can be generated with embeddings across the board (maximum of length 500 even with constraints) while with vocabulary sized parameters, the approach runs of out of memory even without any constraint beyond a length of 20 even on the largest GPU. %This issue becomes even worse when using more than one constraint. On the other hand, \ourmodel is comfortably able work with up to a 1000 tokens without constraint with up to 200 tokens and only fails in a case of a constraint defined on two language models the same size as GPT2-Large, which is caused because three copies of GPT2-Large do not fit into a 24GB GPU.

% This ablation in addition to benefits of defining hard constraints as shown in \Sref{subsec:lexical-control} provides sufficient evidence of advantages of using our presented approach.

% \usepackage{booktabs}
% \usepackage{multirow}
% Limitations: speed can still be improved
% \noindent
% \textbf{Speed and Memory}
% \textbf{Sources of Diversity}
\paragraph{Sources of Diversity}
Our proposed approach has two sources of randomness which can potentially lead to diversity: initialization and noise addition at each step of Langevin Dynamics. To understand their effects, we vary these sources and compute the diversity metrics. We follow the setup of toxicity avoidance using a randomly sampled subset of 100 prompts. The results are shown in \tref{tab:ablations-initialization-threshold}. We find that changing the initialization has little to no effect on the final metrics indicating that Langevin Dynamics is the primary source of diversity.% However [something about schedule]

% \usepackage{booktabs}
% \usepackage{multirow}

% \noindent
% \textbf{Compatibility of Constraints}
\paragraph{Compatibility of Constraints}
Although, our approach allows any combination of constraints in principle, in many cases, the combination might not be compatible. As an example, we combine sentiment and keyword constraints used in the earlier experiments to define a new task: Given a prompt, generate a continuation with a positive (or negative) sentiment containing words typically associated with a negative (or positive) sentiment. Using our best performing constraint (\ourmodel-\textsc{two-disc}) from \Sref{subsec:sentiment-controlled-generation}, and a single keyword constraint, we find that \ourmodel fails almost ${\sim}90\%$ of the times since two constraints are incompatible for most scenarios. For when it does succeed,
we present selected examples in \tref{tab:sentiment-keyword}.

\paragraph{Varying threshold $\epsilon$}
In our experiments, each function $f_i$ is constrained to be bounded by a thresholds $\epsilon_i$, which are tunable hyperparameters. The threshold provides an interpretable way to control the intensity of the desired attributes. To illustrate this capability, we again follow the setup of toxicity avoidance with 100 prompts and apply the constraint $p_\textsc{toxicity} < \epsilon$ with $\epsilon \in \{0.5, 0.3, 0.1, 0.01\}$. As shown in \tref{tab:ablations-initialization-threshold}, making $\epsilon$ smaller improves toxicity control. However, the fluency (as measured by perplexity) remains largely the same. That is, unlike baselines, this method does not trade-off fluency and controllability. However, there is a trade off between diversity and controllability as we observe in sentiment control experiments (\Sref{subsec:sentiment-controlled-generation}) where making a constraint stricter leads to a decline in diversity. 

\section{Conclusion}
We present \ourmodel, a sampling algorithm from language
models that flexibly combines pretrained LMs with any differentiable constraints. Our primary contributions are a (1) gradient based MCMC sampling method (Langevin Dynamics) performed on (2) intermediate representation of tokens (embeddings). With experiments on both soft and hard constraints with different pretrained LMs, we show that this approach generates diverse outputs which better conform both to desired constraints as well as the underlying LM distribution. Despite the observed improvements, we believe we have barely scratched the surface.
% show the usefulness of this approach. 
% In addition to its potential applications in factual rewriting
% and debiasing text, this work holds promise in making language generation models personalizable
% and adaptive to different dialects or even individual speakers, since MUCOCO re-uses pre-trained
% LMs without adaptation and can incorporate constraints (e.g., dialect or user properties) trained on
% 9
% very little data. 
% While we use a simple Langevin Dynamics
% in thi work to facilitate sampling, 
In future work, we will explore ways to improve the convergence properties of this algorithm using more sophisticated MCMC algorithms~\citep{10.2307/41057430} and develop constraints to improve performance on longer sequences. Furthermore, since we perform updates on embeddings rather than vocabulary distributions, future work may also study ways to expand vocabularies at decoding time.  

\section*{Acknowledgements}
The authors would like to thank Xiaochuang Han, Artidoro Pagnoni, Melanie Sclar, Lucille Njoo and Anjalie Field for helpful feedback and discussions, and the
anonymous reviewers for much appreciated feedback.
S.K. is supported by a Google PhD Fellowship.
This material is based upon work supported by the National Science Foundation under CAREER Grant No.~IIS2142739, as well as Grants No.~IIS2125201 and IIS2203097.
The views and opinions
of authors expressed herein do not necessarily state or reflect those of the United States Government
or any agency thereof. 

\section*{Limitations}
% we acknowledge that 
Despite speed improvements on gradient-based decoding compared to using vocabulary-sized representation, this approach still requires iteratively updating $L\times d$ parameters (with each update involve a forward and a backward pass) and is considerably slower than autoregressive decoding methods (anywhere between 15-20 times longer). 
% For example, on a single GeForce RTX 2080 Ti (12GB), with a batch size of 1, and a single classifier constraint, our approach takes approximately 90 minutes on
% average to decode around 1200 sentences compared to around 20 minutes for FUDGE [69] with a
% single constraint
A straightforward way to improve decoding speed is using larger batches and smaller floating point operations which we leave for future work. 
Further improvements may also be achieved by adapting more sophisticated gradient based methods for faster convergence~\citep{10.2307/41057430} or techniques from diffusion models in image generation~\citep{https://doi.org/10.48550/arxiv.2101.02388}.
Like other non-autoregressive decoding approaches, this method also requires pre-defining a fixed output length which can be a hindrance. This is an active area of research with many solutions proposed in the literature including predicting the sequence length~\citep{wang-etal-2021-length}, generating multiple outputs with varying lengths and reranking~\citep{Guo_Tan_He_Qin_Xu_Liu_2019}, continuing generating autoregressively to finish a sentence after a fixed length output is generated~\citep{qin-2022-cold} of all which have shown promising results.
% and even training the LM itself to predict the padding tokens~\citep{} all of which have shown promising results. 
Furthermore, since this algorithm aims to find high probability sequences under the LM, and most open-ended LMs suffer from degeneracy (repeating sequences get high probability)~\citep{Holtzman2020The}, using it can sometimes lead to such generations especially for long sequences (as we observe in \Sref{subsec:sentiment-controlled-generation}). Incorporating repetition reducing loss functions~\citep{Welleck2020Neural,https://doi.org/10.48550/arxiv.2202.00666} as constraints can help alleviate this issue. We leave this exploration for future work.
% Other limitations: predefining a length, defining all constraints as functions of embeddings, repetitions with longer sequences.

\section*{Ethical Considerations}
% Language generation is a growing research area, 
% and state-of-the-art techniques are still not powerful enough to facilitate fine-grained control over generated content. 
Most large LMs trained on web content have been shown to generate harmful language, generating toxic and non-factual content \cite{gehman-etal-2020-realtoxicityprompts,pagnoni2021understanding,sheng-etal-2021-societal,10.1145/3531146.3533088}, with potential malicious use cases~\citep{wallace-etal-2019-universal, wallace-etal-2020-imitation, zellers2019neuralfakenews}. 
Advancements in controlled text generation can be used to mitigate many such biases potentially alleviating these issues
% Controlled text generation techniques can be used to mitigate many such problematic biases already encoded in large language models
~\citep{gehman-etal-2020-realtoxicityprompts,10.1145/3442188.3445922, liu2021onthefly}. They also find useful applications in anonymizing protected attributes in written text~\citep{reddy2016obfuscating}, as writing assistants to avoid users' implicit biases~\citep{ma2020powertransformer, field2020unsupervised}. % However, none of the approaches are perfect.
% They also have many other positive use-cases, for example, anonymizing personal attributes in written text \cite{reddy2016obfuscating}, and even aiding authors in avoiding implicit biases in their writing~\citep{ma2020powertransformer, field2020unsupervised}. However, none of the existing approaches, including ours, can sufficiently address these issues yet.

% While the goal of controlled text generation is to combat these issues, none of the existing techniques, including ours, cannot yet fully address them.
% We warn that these advancements can also be used as a useful tool for mitigating many problematic biases already encoded in large language models~\citep{ gehman2020realtoxicityprompts, 10.1145/3442188.3445922, liu2021onthefly}, for anonymizing personal attributes \cite{reddy2016obfuscating}, and even as aids for humans to avoid implicit biases in their writing~\citep{ma2020powertransformer, field2020unsupervised}. 
A risk with controllability is that it can be used with mal-intent 
to generate misinformation, make output text more biased and toxic as well as target individuals to influence public opinion. To combat these issues, apart from controllable generation, future research should focus on developing better defense methods against misusing these models, in a way that could cause societal harms~\citep{Kumar2022LanguageGM}.
% For example, when style transfer techniques are used in conjunction with users' personal attributes such as gender, they can amplify harmful social biases. We thus opted not to include gender transfer in our experiments.  %Our goal in this work is to enable finer-grained control over generated texts that could potentially alleviate these issues. 

% Nevertheless, these issues should not discourage the scientific exploration that will advance the state-of-the-art in many positive usages of controlled text generation, including in machine translation, question answering, summarization, dialogue, etc. 
% In parallel, future research should focus on developing better defense methods against mis-using these models maliciously, in a way that could cause societal harms~\citep{zellers2019neuralfakenews}.   


%%%TODO ack: This material is based upon work supported by the National Science Foundation under Grants No.~IIS2125201, IIS2007960, and IIS2040926.
\bibliography{anthology,custom}
\bibliographystyle{acl_natbib}

\appendix

\section{\ourmodel Decoding Algorithm}
We provide a %figure representing our method in~\fref{fig:mucola} with details of constraints in \fref{fig:primary-objective} and a 
formal algorithm for \ourmodel in \ref{algo}.
\begin{algorithm}
\SetAlgoLined
\textbf{Input:} input sequence $\mathbf{x}$, output length $L$, base LM, attribute functions $f_i$ and their respective thresholds $\epsilon_i$, step sizes $\eta$, $\eta_{max}$ (and schedule), $\eta_\lambda$, initial noise variance $\beta_\text{init}$ (and schedule)\; 
\KwResult{output sequence $\mathbf{y}$}
 For all $n \in \{1, \ldots, L\}$, initialize $\tilde{e}_n^{(0)}$\; 
 For all $i \in \{1, \ldots u\}$, initialize $\lambda_i^{(0)}$ as $0$\;
 Initialize $\beta^{(0)}$ as $\beta_\text{init}$\;
 Initialize $\eta^{(0)}$ as $\eta$\;
 \For{$t = 1, \ldots, \textsc{MaxSteps}$}{
 \tcp{forward pass}
  compute the energy function $\mathcal{E}$ (see \Sref{subsec:lagrangian})\;
  \tcp{backward pass}
  for all $n, i$, compute $\nabla_{\tilde{e}_n}^{(t-1)} = \frac{\partial \mathcal{E}}{\partial \tilde{\mathbf{e}}_n}$, $\nabla_{\lambda_i}^{(t-1)} = \frac{\partial \mathcal{E}}{\partial \lambda_i}$\; %\algorithmiccomment{This is a comment}
  \tcp{Update the parameters}
  Sample $z^{(t-1)} \sim \mathcal{N}(0, I_d)$\;
  Update $\mathbf{\tilde{e}_{y}}^{t} = \mathbf{\mathrm{Proj}}_\mathbf{E}(\mathbf{\tilde{e}_{y}}^{(t-1)} - \eta \nabla_{\mathbf{\tilde{e}_y}}^{(t-1)} \mathcal{E}  + \sqrt{2\eta^{(t-1)}\beta}z^{(t-1)})$\;
%   update $\tilde{e}_n^{(t)} = \tilde{e}_n^{(t-1)} \exp (\eta_1 \nabla_{\tilde{e}_n} \mathcal{E})$\;
  Update $\lambda_i^{t} = \max(0, \lambda_i^{t-1} + \eta_2 \nabla_{\lambda_i}^{(t-1)} \mathcal{E}$)\;%, and $\mu_i^{t} = \max(0, \mu_i^{t-1} + \eta_2 \nabla_{\mu_i} \mathcal{L})$\;
  update $\beta^{(t)}, \eta^{(t)}$ following the threshold update schedule.
 }
Convert $\mathbf{\tilde{e}}^{(t)}$ to discete tokens $\mathbf{\hat{y}}^{(t)}$ by nearest neighbor search.\;
\textbf{return} $\mathrm{arg}\min_t \{-\log P(\hat{\mathbf{y}}^{(t)}|\mathbf{x}): \forall i, f_i(\tilde{\mathbf{y}}^{(t)} | [\mathbf{x}]) \leq \epsilon_i\}$;
 \caption{\ourmodel: detailed decoding algorithm}
 \label{algo}
\end{algorithm}

\section{Related Work}

\paragraph{Controllable Text Generation}
Prior work in this area can be divided into three categories: The first focuses on training models with specific control codes via pretraining~\citep{keskarCTRL2019} or finetuning~\citep{gururangan-etal-2020-dont,chan2021cocon} for prompt based generation, and generative models for tasks such for style transfer~\citep{subramanian2019multipleattribute, ziegler2020finetuning, prabhumoye2018style, yu2017seqgan}. These methods are naturally difficult to extend to new controls as it requires retraining the models. 

The second category includes decoding approaches from LMs without modifying them (\ourmodel falls under this category). Most prior work in this space has explored methods to modify left-to-right search or sampling algorithms by modifying the output probability distribution at each step using different control functions. \citet{Dathathri2020Plug,KrauseGeDi2020,yang2021fudge,liu-etal-2021-dexperts} apply this approach for soft constraints defined by classifiers and LMs whereas \citet{lu-etal-2021-neurologic,Lu2021NeuroLogicAD,pascual-etal-2021-plug-play} develop heuristic control functions for keyword based constraints. In contrast, we show that \ourmodel is able to incorporate both kinds of constraints. Since these approaches generate one token at time and do not allow modifying a token once it is generated, they are not ideal for controls that are conceptually defined on the entire sequence. Hence, prior work has also explored non-autoregressive decoding methods~\citep{mix-and-match-2022}. Most closely related to \ourmodel is \citet{kumar2021controlled} which propose a gradient-based decoding algorithm which we extend to generate multiple samples. Also related is \citet{qin-2022-cold} that perform Langevin Dynamics in the simplex space to incorporate control by representing the energy function as a linear combination of control functions. In contrast, we represent the energy functions as a Lagrangian and perform these updates on a much smaller embedding space allowing us to generate longer sequences.

The third category includes more recent few-shot methods which rely on prompting large LMs such as GPT3 to incorporate controls based on demonstrations~\citep{qian-etal-2022-controllable,https://doi.org/10.48550/arxiv.2204.13362,carlsson-etal-2022-fine}. \ourmodel is an orthogonal approach to this work and can be applied on top of prompt-based solutions to increase control satisfaction.

While this work and the work described above focuses on controlling attributes of individual text outputs, in related work, \citet{khalifa2021a,pmlr-v162-korbak22a} develop approaches for distributional control of various attributes in generated text corpora.

\paragraph{Gradient-based Sampling}
Langevin Dynamics and other gradient-based MCMC methods have been developed for generative modeling in continuous domains such as images~\citep{NEURIPS2019_3001ef25} and audio~\citep{pmlr-v139-jayaram21b} among others where the models are trained to predict the gradients (via a score function) directly whereas \ourmodel requires a backward pass to compute them. Also related are diffusion models which have obtained state-of-the-art performance for many generative tasks~\citep{ramesh2022hierarchical,ho2022video}. Similar ideas have also been applied to train text generation models in concurrent work with promising results for incorporating controls~\citep{Li-2022-DiffusionLM}. 

\section{Implementation Details}
\label{appsec:implementation-details}
Here, we describe additional implementation details as part of the experimental setup described in \Sref{sec:experiments}

\paragraph{Noise Schedule}
The amount of noise in each update is controlled by $\beta$ (\Sref{subsec:langevin}) which represents the variance of the noise term. We initialize $\beta$ with $5.0$ and decrease it to $0.05$ in a geometric progression for $100$ steps after which we keep it constant at $0.05$ for the remaining $150$ steps. The range of $\beta$ is guided by best practices in~\citet{NEURIPS2019_3001ef25} prescribing the initial variance to be close to the maximum distance between any two vectors in the input space and the minimum value being close to $0$. This schedule allows for sufficient exploration in the beginning helping in diversity, while, leaving enough iterations for optimizing the final output into a fluent sequence.

\paragraph{Step Size and Selection Criterion}
The step-size $\eta$ in projected gradient descent depends on the geometry of the embedding space of the underlying language model. Since we project back the update at every step to $\mathbf{E}$, if the update term is not big enough in the projected gradient update, the sequence at step $t+1$ would remain the same. This observation provides a very simple criterion for early stopping and selecting the best output out of all iterations. When the additive noise is small (near the end of optimization), the update term can be small due to following factors: (a) $\eta$ is small, (b) the gradient $\nabla \mathcal{E}_\mathbf{\tilde{e}}$ is small which implies the output sequence has ``converged''. Hence, we define a schedule on the step-size as follows: we start with a step-size $\eta$, and update the outputs using Langevin Dynamics until the sequence stops updating, i.e., the update value becomes too small (and satisfies all constraints). Now, to make sure that this is a convergence point and not a result of the step size being too small, we update the step size linearly to $\eta_\textrm{max}$ in $s$ steps\footnote{$s$ is empirically defined as $40$ in all our experiments.}. If the sequence does not update in $s$ steps, we stop early and predict the output. Otherwise, the process continues. If it does not stop early at the end of maximum number of steps, we predict the output with the highest likelihood which repeated at least $5$ times. In the event, no such repetition is observed, we deem the optimization as ``failed'' and restart. If the restarts also fail, we just predict the autoregressive output (which in our experiments is obtained with nucleus sampling with $p=0.96$). This fallback mechanism ensures that the output, irrespective of the constraint satisfaction is always a sample of $P$ while preventing generating half-baked outputs.

% We conduct experiments with GPT2-Large and GPT2-XL

\paragraph{Multipliers Update Schedule}
We initialize each of the multipliers $\lambda_i$ with 0, update the multipliers via gradient ascent every $20$ steps using the step-size $1.0$. In addition, if the sequence stops updating at a certain iteration (as described above) and $i$-th constraint is not satisfied, we update $\lambda_i$ at every iteration till the sequence starts updating again. This schedule prevents fluctuation in the multiplier values when the noise is high in the early iterations and the sequence has not converged to anything fluent while still allowing updates when required~\citep{NIPS1987_a87ff679,paria2020minimizing}.

\section{Training Details for Soft Constraint Models}
\label{appsec:training-details}
Since we decode by computing gradients over token embeddings, it requires that all constraint models share the same embedding table $\mathbf{E}$ as that of the underlying language model $P$. Since any typical text based model involves an embedding table, we can train a constraint using such a model
% We train each constraint model $f_i$, be it a classifier or a language model, 
by simply initializing its embedding table with $\mathbf{E}$. In principle, this initialization allows using any off-the-shelf pretrained model as a constraint function by finetuning it on appropriate data. In our experiments, we use the following models in different experiments:
\paragraph{Toxicity Classifier}
For toxicity avoidance (\Sref{subsec:toxicity}), we finetune \href{https://huggingface.co/roberta-base}{roberta-base}~\citep{liu2019roberta} with a binary classification head using a dataset of human-annotated comments from the Jigsaw Unintended Bias In Toxicity Classification Kaggle Challenge. The dataset has ${\sim}160K$ toxic comments and ${\sim}1.4M$ non-toxic comments. We first balance this dataset by subsampling $160K$ examples from the non-toxic class. We replace the embedding table of roberta-base with that of the underlying LM (GPT2-Large in our case). To address the dimension mismatch of the two embedding tables, during finetuning, we also learn a linear projection matrix which transforms base LM embedding to a smaller dimension of roberta-base. We keep base LM embedding frozen during this finetuning. We use a learning rate of $1e-5$ and train for 3 epochs with an effective batch size of 64. We choose a checkpoint with an accuracy of ${\sim}93\%$ on a heldout development set.

\paragraph{Sentiment Classifiers}
For sentiment control experiments in \Sref{subsec:sentiment-controlled-generation}, we experiment with different kinds of constraints defined using both classifiers and language models. For both setups we use two datasets: SST-2 corpus~\citep{socher-etal-2013-recursive} containing ${\sim}4$K examples in Movie reviews for each class; and Yelp polarity corpus containing ${\sim}280$K examples for each class containing a mixed domain of reviews. 

For discriminative classifiers, we also finetune roberta-base using the same setup and hyperparameters as the toxicity classifier. Our best model obtains an accuracy of ${\sim} 92\%$ on the SST-2 test set and ${\sim} 98\%$ on the Yelp test set.

To train the generative classifiers, we finetune GPT2-Large (and do not need to substitute any embedding tables) keeping the embedding table frozen. We use the loss $-\log p_\text{gen}(\text{label}|x)$ for each training instance where $p_\text{gen}(\text{label}=0|\text{text}) = p_\text{LM}(\text{text}|\text{label}=0)/(p_\text{LM}(\text{text}|\text{label}=0) + p_\text{LM}(\text{text}|\text{label}=1))$.
This is due to Bayes' rule ($p(\text{label})$ vanishes as we set it to 0.5 for balanced datasets). Here $p_\text{LM}(\text{text}|\text{label})$ is obtained using the language model by computing the probability of the text conditioned on the input token ``positive'' for the positive label and ``negative'' otherwise. 

We again follow the same training hyperparameters for this setup. On SST-2 test set, we obtain an accuracy of ${\sim}95\%$ and on Yelp, we obtain an accuracy of ${\sim}98\%$.

\section{Additional Explanation and Results for Hard Constraint}
\label{appsec:keyword-constraint}
The keyword distance function $d(w, \tilde{\mathbf{e}})$ is computed in three steps. First, we convert each $\tilde{e}_n$ to a ``probability'' over the vocabulary as,
\begin{align*}
    \pi_n = \mathrm{softmax}(-\|\tilde{e}_n - e_1\|^2_2, \ldots, -\|\tilde{e}_n - e_{|\mathcal{V}|}\|^2_2)
    % \label{eq:l2softmax}
\end{align*}
where $\{e_1, \ldots, e_{|\mathcal{V}|}\}$ are entries in the embedding table $\mathbf{E}$. Since each $\tilde{e}_n$ itself also corresponds to a vector in $\mathbf{E}$, if $n$-th token in the sequence is $w$, then, $\|\tilde{e}_n - e_w\|^2_2$ would be 0 leading to 
% say corresponding to the token index $v$, this transformation assigns $\pi_{n, v}$ the highest value in $\pi_n$, since the corresponding Euclidean distance would be $0$ and others negative.
% In other words, if $k$-th token in the sequence is $w$, then, 
$\pi_{n, w} = \max_j \pi_{n, j}$.
% our goal in defining this similarity function is to have $w$'s embedding $e_w$ as one of the vectors in the output sequence $\mathbf{\tilde{e}}$ or in other words, for which $\pi_{k, w}$ is the maximum value in $\pi_k$.
That is, maximizing $g_n = \log \pi_{n, w}$ with respect to $\tilde{e}_n$ would nudge it towards the $e_w$. 
Since, we don't know which index we want $w$ to appear at in advance, following~\citep{editinvariant2022,qin-2022-cold}, we (soft) sample it using $\pi_{n, w}$ as weights. This brings us to the second step, as we define, $q = \textsc{gumbel-softmax}(-g_1/\tau, \ldots, -g_N/\tau)$
where $\tau$ is the temperature. We use hard sampling here to ensure $q$ is one-hot. Finally, we define the constraint function as, $d(w, \mathbf{\tilde{e}}) = \sum_{i=n}^N -q_n g_n$. Intuitively, this function aims to generate the word $w$ wherever it already has a high chance of getting generated (measured via $\pi_{n, w}$'s). Stochasticity in this function allows for exploration.
This function can be easily extended from words to phrases of length $l$, $w = (w_1, \ldots, w_l)$ by defining $g_n = \frac{1}{l}\sum_{u=1}^l -\log \pi_{w_u, n+u}$.
% $l$ distances and averaging them to compute each $g_n$, as $g_n = \sum_{u=1}^l \|e_{w_u} - \tilde{e}_{n+u}\|$. In other words, we consider every $l$ length subsequence in the output $\mathbf{\tilde{e}}$ and compute $g_n$ as the average of its Euclidean distance from corresponding tokens in the phrase $w$. 
This computation can be efficiently done on a GPU using a convolution operation~\citep{editinvariant2022}.

% This function can be easily extended from words to phrases of length $l$, $w = (w_1, \ldots, w_l)$ by defining $l$ distances and averaging them to compute each $g_n$, as $g_n = \sum_{u=1}^l \|e_{w_u} - \tilde{e}_{n+u}\|$. In other words, we consider every $l$ length subsequence in the output $\mathbf{\tilde{e}}$ and compute $g_n$ as the average of its Euclidean distance from corresponding tokens in the phrase $w$. This computation can be efficiently done in on a GPU using a convolution operation~\citep{editinvariant2022}.

Based on this definition, we define the keyword constraint for \ourmodel as $d(w, \mathbf{\tilde{e}}) \leq -\log \pi_{w, w} + \delta$, where $\delta$ is a small positive value (we set it as 0.1). $\pi_w$ is a slight abuse of notation to define a distribution similar to $\pi_n$ ($n$ refers in an index in sequence whereas $w$ refers to an index in $\mathcal{V}$). Note that the threshold for each keyword is different.\footnote{While we do not experiment with it in this work, the constraint $K(w, \mathbf{\tilde{e}})$ can be easily extended to setup where at least one out of $n$ given words (for example different surface forms of the same root), $S = \{w_1, \ldots, w_p\}$ must appear in the output by defining a new constraint as $K(S, \mathbf{\tilde{e}}) = \max_{w_i \in S} K(w_i, \mathbf{\tilde{e}})$ or its soft version using the gumbel-softmax trick.} 

Intuitively, if $w$ appears in the output at the $k$-th position, then $\pi_{k, w} = \pi_{w, w} = \max_j \pi_{k, j}$ with $q_k$ as $1$. This reduces the distance function to $-\log \pi_{k, w}$ which is less than the defined threshold. 
Conversely, if $w$ does not appear in the output, for each $n$, $-\log \pi_{n, w}$ would be higher than $\log \pi_{w, w}$ and the constraint remains unsatisfied. This is due to an empirical observation we make in all embedding tables we use, that $\pi_{w,w} = \max_j \pi_{w, j} = \max_j \pi_{j, w}$. In other words, not only is the probability of a word under its own distribution $pi_w$ the greater than probability of all other words (since the corresponding distance is 0), it is also larger than $w$'s probability under all other distributions defined for any word in the vocabulary. Under the assumption that minimum distance between any two vectors in the table is greater than a small positive value, we conjecture this claim to be true for any embedding table.

\paragraph{Open-Ended Keyword Guided Generation}
In addition to \textsc{CommonGen}, we report results on \textsc{ROC}~\citep{pascual-etal-2021-plug-play} task where given 5 keywords, the goal is generate a sequence of max length $90$ containing those terms. For both datasets, for set of keywords, we generate samples of length 10, 20, and 40 (with 3 restarts for each) and after all iterations are complete, we continue generating more tokens autoregressively until a maximum of $40$
($90$ in case of \textsc{ROC}) 
tokens are generated or end of sequence token is generated. 
Finally, we evaluate on one output which satisfies the constraints and has the lowest perplexity according to the LM.
% For each set of keywords, we generate 4 samples, and select the one which satisfies the constraints and has the lowest perplexity.
% For both datasets, 
We compare \ourmodel with the best reported results in~\citep{qin-2022-cold} and ~\citep{pascual-etal-2021-plug-play} and corresponding baselines. The results for \textsc{ROC} can be found in \tref{tab:roc-results}.

\paragraph{Terminology Constrained Translation}
We follow the setup in~\citet{dinu-etal-2019-training} and use an off-the-shelf English to German translation model by MarianMT~\citep{junczys-dowmunt-etal-2018-marian} to translate a subset of WMT17 en-de test set~\citep{bojar-etal-2017-findings}. The constraint here is to integrate a given custom terminology into the translation output; where the terms are automatically created from the IATE EU terminology database for 414 test sentences (with 1 to 3 terminology constraint per example). We use \citet{Lu2021NeuroLogicAD} as our best baseline and also report other baselines reported by them. We generate each translation by first generating with beam search unconstrained (with beam size of 6). If this output is of length $L$. We use \ourmodel to generate sequences of length $\{L, L+1, \ldots, L+10\}$ and select the generation which has the highest length-normalized log-probability as the final translation. We evaluate on BLEU score\footnote{For fair comparison, we compute a tokenized BLEU score reported by the baselines following \url{https://github.com/INK-USC/CommonGen/tree/master/evaluation}} and coverage accuracy. As reported in \tref{tab:translation-keyword}, \ourmodel obtains perfect (100\%) coverage while at the same maintaining BLEU score. 

% \usepackage{booktabs}
\begin{table}[]
\centering
\begin{tabular}{@{}lll@{}}
\toprule
\textbf{Method} & \textbf{BLEU} & \textbf{Coverage} \\ \midrule
Unconstrained & 32.9 & 85.3 \\
\citet{post-vilar-2018-fast} & 33.0 & 94.3 \\
Neurologic* & 33.5 & 97.2 \\
\ourmodel & 33.1 & 100 \\ \bottomrule
\end{tabular}
\caption{Results for terminology constrained en--de translatoin (\Sref{appsec:keyword-constraint})}
\label{tab:translation-keyword}
\end{table}

\paragraph{Entity Constrained Summarization}
 In this setup, we do a preliminary exploration on text summarization with a constraint that a specific entity must appear in the summary given the article. We use BART-Large~\citep{lewis-etal-2020-bart} finetuned on the CNN/Dailymail Corpus~\citep{see-etal-2017-get} as our underlying LM. First, we obtain all named entities appearing in the article using an off-the-shelf recognizer\footnote{https://huggingface.co/dslim/bert-base-NER-uncased}. We then use \ourmodel to sample a summary (of maximum length 50) from the model considering appearance of each entity as a constraint. We show selected examples with promising results in \tref{tab:summarization-examples}, \tref{tab:summarization-examples2} and \tref{tab:summarization-examples3}. Evaluating this setup is non-trivial, since it adds new sentences/phrases to the summary and will naturally perform poorly on standard reference based metrics such as ROUGE. Hence, we leave this evaluation for future work.

\section{Additional Results for Soft Constraints}
\label{appsec:additional-results}

\paragraph{Toxicity Avoidance}
For human evaluation, we follow an A/B testing framework and compare \ourmodel and DExperts. We sample 200 prompts from the test set and consider 2 generations per prompt. We ask each annotator to rank the outputs from the two approaches on (1) toxicity if one output is more or less toxic than the other, or if both are equally toxic/non-toxic, (2) topicality: is the generation coherent with the prompt and follows the same general topic, and (3) fluency: if the outputs have any grammatical mistakes. We collect 3 annotations per pair. We find that in terms of toxicity, both models perform similarly with an average 8.5\% annotations preferring \ourmodel's outputs compared to 9.5\% for DExperts (rest are equally ranked). On topicality, 22.5\% of annotations prefer \ourmodel's outputs while 19\% prefer Dexperts (rest are equally ranked). On fluency, both models perform similarly with 22.5\% and 23\% in each method's favor and rest equally ranked.

\paragraph{Sentiment Control}
We present the full set of results for sentiment control experiments in tables~\ref{tab:sentiment-results-12-positive},
%\ref{tab:sentiment-results-12-negative}, 
\ref{tab:sentiment-results-20-positive}, 
%\ref{tab:sentiment-results-20-negative}, 
\ref{tab:sentiment-results-50-positive} and %\ref{tab:sentiment-results-50-negative}. 
More details can be found in the captions. For human evaluation, we similarly follow an A/B testing framework and compare \ourmodel and DExperts (for outputs of length 20). We consider all 15 prompts from the test set and consider 2 generations per prompt. We ask each annotator to rank the outputs from the two approaches on (1) positivity if one output is positive and the other is not, or if both are positive/not-positive, (2) topicality: is the generation coherent with the prompt and follows the same general topic, and (3) fluency: if the outputs have any grammatical mistakes. We collect 3 annotations per pair. We find that in terms of positivity, on an average 23.3\% annotations prefer \ourmodel's outputs compared to 16.7\% for DExperts (rest are equally ranked). On topicality, 26.7\% of annotations prefer \ourmodel's outputs while 13.3\% prefer Dexperts (rest are equally ranked). On fluency, \ourmodel slightly underperforms with 7.8\% and 10\% in each method's favor and rest equally ranked.

\begin{table*}
    \centering
    \footnotesize
    \begin{tabular}{@{}llcccccccc@{}}
    \toprule
    \multirow{2}{*}{\textbf{Approach}} &
    \multirow{2}{*}{\textbf{Setting}} &
    \multicolumn{3}{c}{\textbf{\% Positive Sentiment ($\downarrow$) }} & \multicolumn{2}{c}{\textbf{Fluency}}& \multicolumn{3}{c}{\textbf{Diversity}} \\ \cmidrule(l){3-10} 
    & & \multicolumn{1}{c}{\textsc{c1}}& \multicolumn{1}{c}{\textsc{c2}}& \multicolumn{1}{c}{\textsc{c3}}& \multicolumn{1}{c}{Perplexity} & \multicolumn{1}{c}{\begin{tabular}[c]{@{}c@{}}CoLa\\Accuracy\end{tabular}} & \multicolumn{1}{c}{Dist-1} & \multicolumn{1}{c}{Dist-2} & \multicolumn{1}{c}{Dist-3} \\ \midrule
\textsc{GPT-2} & - & 49.0 & 45.0 & 62.0 & 54.9 & 68.7 & 0.66 & 0.87 & 0.81 \\ \midrule
\textsc{DAPT}  & SST-2 & 71.3 & 66.7 & 75.0 & 98.0 & 64.0 & 0.64 & 0.85 & 0.79 \\
\textsc{DAPT}  & Yelp & 64.0 & 71.3 & 79.7 & 146.6 & 58.0 & 0.60 & 0.84 & 0.80 \\ \midrule
\textsc{FUDGE} & SST-2 & 71.7 & 70.0 & 79.0 & 11.4 & 82.7 & 0.53 & 0.76 & 0.77\\
\textsc{FUDGE} & Yelp & 71.7 & 73.7 & 84.7 & 11.8 & 85.7 & 0.53 & 0.76 & 0.77 \\ 
\ourmodel-\textsc{disc} & SST-2 & 90.0 & 81.7 & 93.3 & 28.8 & 67.3 & 0.52 & 0.73 & 0.74 \\
\ourmodel-\textsc{disc} & Yelp & 88.3 & 87.0 & 91.7 & 32.9 & 64.3 & 0.52 & 0.74 & 0.75 \\ 
\ourmodel-\textsc{two-disc} & Yelp, SST2 & 94.0 & 91.3 & 94.7 & 29.4 & 55.0 & 0.46 & 0.68 & 0.71 \\\midrule
\textsc{GeDi}  & SST-2 & 99.7 & 91.0 & 99.3 & 625.7 & 54.3 & 0.65 & 0.76 & 0.71 \\
\textsc{GeDi}  & Yelp & 82.0 & 90.0 & 89.0 & 444.9 & 40.0 & 0.71 & 0.78 & 0.66 \\
\ourmodel-\textsc{gen}  & SST-2 & 91.3 & 88.3 & 97.0 & 57.2 & 68.0 & 0.50 & 0.69 & 0.70 \\
\ourmodel-\textsc{gen}  & Yelp & 86.3 & 89.7 & 91.7 & 53.0 & 67.7 & 0.50 & 0.70 & 0.70  \\ 
\ourmodel-\textsc{prompt}  & - &  89.0 & 88.7 & 94.7 & 43.7 & 66.7 & 0.49 & 0.72 & 0.73 \\
\midrule
\textsc{DExperts} & SST-2 & 93.1 & 86.9 & 94.9 & 75.2 & 71.5 & 0.63 & 0.85 & 0.81 \\
\textsc{DExperts} & Yelp & 80.3 & 88.5 & 88.8 & 116.3 & 67.5 & 0.67 & 0.84 & 0.79 \\
\ourmodel-\textsc{DExperts}  & SST-2 & 93.0 & 88.0 & 94.0 & 41.4 & 66.3 & 0.47 & 0.71 & 0.73  \\
\ourmodel-\textsc{DExperts}  & Yelp & 74.3 & 74.0 & 83.3 & 72.5 & 66.0 & 0.52 & 0.73 & 0.74  \\ 
\bottomrule
\end{tabular}
\caption{Positive sentiment control results on outputs of length 12. For each baseline (\textsc{FUDGE}, \textsc{GeDi} and \textsc{DExperts}), we convert their respective constraints to a classifier (generative or discriminative; see \Sref{subsec:sentiment-controlled-generation}). For \textsc{FUDGE} and \textsc{GeDi}, we show improvements on both control (\% positive sentiment) and fluency (Perplexity) without any model specific changes. This improvement is consistent on models trained on both datasets (SST-2 and Yelp). \textsc{DExperts} outperforms all baselines here including our method.}
\label{tab:sentiment-results-12-positive}
\end{table*}

\begin{table*}
    \centering
    \footnotesize
    \begin{tabular}{@{}llcccccccc@{}}
    \toprule
    \multirow{2}{*}{\textbf{Approach}} &
    \multirow{2}{*}{\textbf{Setting}} &
    \multicolumn{3}{c}{\textbf{\% Positive Sentiment ($\uparrow$) }} & \multicolumn{2}{c}{\textbf{Fluency}}& \multicolumn{3}{c}{\textbf{Diversity}} \\ \cmidrule(l){3-10} 
    & & \multicolumn{1}{c}{\textsc{c1}}& \multicolumn{1}{c}{\textsc{c2}}& \multicolumn{1}{c}{\textsc{c3}}& \multicolumn{1}{c}{Perplexity} & \multicolumn{1}{c}{\begin{tabular}[c]{@{}c@{}}CoLa\\Accuracy\end{tabular}} & \multicolumn{1}{c}{Dist-1} & \multicolumn{1}{c}{Dist-2} & \multicolumn{1}{c}{Dist-3} \\ \midrule
\textsc{GPT-2} & - & 46.7 & 47.7 & 61.3 & 38.6 & 78.7 & 0.64 & 0.90 & 0.88 \\ \midrule
\textsc{DAPT}  & SST-2 & 73.6 & 70.0 & 78.3 & 76.9 & 70.7 & 0.64 & 0.89 & 0.86 \\
\textsc{DAPT}  & Yelp & 65.0 & 75.0 & 80.7 & 86.6 & 69.7 & 0.59 & 0.88 & 0.87 \\ \midrule
\textsc{FUDGE} & SST-2 & 67.6 & 63.0 & 79.3 & 10.3 & 94.0 & 0.51 & 0.80 & 0.84 \\
\textsc{FUDGE} & Yelp & 71.0 & 70.0 & 79.3 & 10.6 & 89.0 & 0.53 & 0.81 & 0.85  \\ 
\ourmodel-\textsc{disc} & SST-2 & 84.6 & 77.5 & 88.0 & 27.9 & 80.8 & 0.50 & 0.81 & 0.82 \\
\ourmodel-\textsc{disc} & Yelp & 83.0 & 83.6 & 83.0 & \textbf{32.2} & 76.0 & 0.50 & 0.75 & 0.80 \\ 
\ourmodel-\textsc{two-disc} & Yelp, SST2 & \textbf{93.7} & \textbf{91.0} & \textbf{96.0} & 28.9 & 76.7 & 0.53 & 0.77 & 0.74 \\\midrule
\textsc{GeDi}  & SST-2 & 99.0 & 96.3 & 99.7 & 268.7 & 54.0 & 0.69 & 0.87 & 0.84 \\
\textsc{GeDi}  & Yelp & 84.0 & 95.7 & 91.0 & 208.3 & 44.0 & 0.76 & 0.87 & 0.81 \\
\ourmodel-\textsc{gen}  & SST-2 & 86.3 & 80.3 & 93.3 & 45.6 & 77.7 & 0.50 & 0.74 & 0.78 \\
\ourmodel-\textsc{gen}  & Yelp & 79.7 & 83.0 & 90.0 & 27.2 & 72.3 & 0.50 & 0.82 & 0.86  \\ 
\ourmodel-\textsc{prompt}  & - & 87.3 & 91.0 & 93.0 & 53.0 & 77.2 & 0.54 & 0.82 & 0.80  \\
\midrule
\textsc{DExperts} & SST-2 & 91.2 & 83.4 & 95.4 & 55.37 & 81.6 & 0.61 & 0.89 & 0.87 \\
\textsc{DExperts} & Yelp & 81.1 & 85.8 & 92.5 & 95.87 & 71.7 & 0.66 & 0.89 & 0.87 \\
\ourmodel-\textsc{DExperts}  & SST-2 & 89.3 & 83.7 & 93.7 & \textbf{32.2} & 79.7 & 0.51 & 0.78 & 0.80  \\
\ourmodel-\textsc{DExperts} & Yelp & 78.0 & 75.7 & 83.3 & 34.1 & 68.3 & 0.52 & 0.77 & 0.81  \\
\bottomrule
\end{tabular}
\caption{Positive sentiment control results on outputs of length 20. For each baseline (\textsc{FUDGE}, \textsc{GeDi} and \textsc{DExperts}), we convert their respective constraints to a classifier (generative or discriminative; see \Sref{subsec:sentiment-controlled-generation}). For \textsc{FUDGE} and \textsc{GeDi}, we show improvements on both control (\%positive sentiment) and fluency (Perplexity) without any model specific changes. This improvement is consistent on models trained on both datasets (SST-2 and Yelp).}
\label{tab:sentiment-results-20-positive}
\end{table*}

\begin{table*}
    \centering
    \footnotesize
    \begin{tabular}{@{}llcccccccc@{}}
    \toprule
    \multirow{2}{*}{\textbf{Approach}} &
    \multirow{2}{*}{\textbf{Setting}} &
    \multicolumn{3}{c}{\textbf{\% Positive Sentiment ($\downarrow$) }} & \multicolumn{2}{c}{\textbf{Fluency}}& \multicolumn{3}{c}{\textbf{Diversity}} \\ \cmidrule(l){3-10} 
    & & \multicolumn{1}{c}{\textsc{c1}}& \multicolumn{1}{c}{\textsc{c2}}& \multicolumn{1}{c}{\textsc{c3}}& \multicolumn{1}{c}{Perplexity} & \multicolumn{1}{c}{\begin{tabular}[c]{@{}c@{}}CoLa\\Accuracy\end{tabular}} & \multicolumn{1}{c}{Dist-1} & \multicolumn{1}{c}{Dist-2} & \multicolumn{1}{c}{Dist-3} \\ \midrule
\textsc{GPT-2} & - & 47.7 & 44.3 & 61.3 & 36.3 & 78.3 & 0.59 & 0.92 & 0.94 \\ \midrule
\textsc{DAPT}  & SST-2 & 93.0 & 84.3 & 91.7 & 55.3 & 88.0 & 0.61 & 0.92 & 0.94 \\
\textsc{DAPT}  & Yelp & 72.3 & 80.7 & 85.0 & 46.1 & 84.3 & 0.51 & 0.90 & 0.94 \\ \midrule
\textsc{FUDGE} & SST-2 & 71.0 & 61.3 & 84.7 & 8.5 & 98.3 & 0.47 & 0.83 & 0.92 \\
\textsc{FUDGE} & Yelp & 72.3 & 68.0 & 80.3 & 8.3 & 99.0 & 0.47 & 0.83 & 0.92 \\ 
\ourmodel-\textsc{disc} & SST-2 & 88.7 & 81.0 & 91.3 & 15.3 & 72.7 & 0.42 & 0.68 & 0.76\\
\ourmodel-\textsc{disc} & Yelp & 70.7 & 74.3 & 81.3 & 19.1 & 77.7 & 0.48 & 0.77 & 0.85\\ 
\ourmodel-\textsc{two-disc} & Yelp, SST2 & 94.0 & 91.3 & 94.7 & 29.4 & 75.0 & 0.57 & 0.78 & 0.79 \\\midrule
\textsc{GeDi}  & SST-2 & 86.7 & 98.7 & 96.7 & 148.4 & 68.3 & 0.75 & 0.94 & 0.93 \\
\textsc{GeDi}  & Yelp & 99.7 & 98.7 & 100.0 & 114.5 & 74.3 & 0.66 & 0.93 & 0.93 \\
\ourmodel-\textsc{gen}  & SST-2 & 85.0 & 76.3 & 91.0 & 22.5 & 63.7 & 0.44 & 0.71 & 0.78 \\
\ourmodel-\textsc{gen}  & Yelp & 77.7 & 80.7 & 88.3 & 23.4 & 65.0 & 0.43 & 0.69 & 0.76 \\ 
\ourmodel-\textsc{prompt}  & - & 81.3 & 83.0 & 92.7 & 18.2 & 72.0 & 0.39 & 0.67 & 0.77 \\
\midrule
\textsc{DExperts} & SST-2 & 98.1 & 92.0 & 99.5 & 39.5 & 88.5 & 0.57 & 0.91 & 0.94\\
\textsc{DExperts} & Yelp & 87.2 & 91.7 & 94.9 & 54.0 & 77.3 & 0.62 & 0.92 & 0.93\\
\ourmodel-\textsc{DExperts}  & SST-2 & 72.7 & 71.7 & 84.7 & 28.2 & 69.0 & 0.45 & 0.75 & 0.83\\
\ourmodel-\textsc{DExperts}  & Yelp & 62.3 & 61.7 & 75.7 & 18.8 & 81.0 & 0.48 & 0.77 & 0.83\\ 
\bottomrule
\end{tabular}
\caption{Positive sentiment control results on outputs of length 50. For each baseline (\textsc{FUDGE}, \textsc{GeDi} and \textsc{DExperts}), we convert their respective constraints to a classifier (generative or discriminative; see \Sref{subsec:sentiment-controlled-generation}). For \textsc{FUDGE} and \textsc{GeDi}, we show improvements on both control (\% positive sentiment) and fluency (Perplexity) without any model specific changes. This improvement is consistent on models trained on both datasets (SST-2 and Yelp).}
\label{tab:sentiment-results-50-positive}
\end{table*}

\begin{table*}[]
\begin{tabular}{@{}lllllllll@{}}
\toprule
\multicolumn{1}{l}{\multirow{2}{*}{\textbf{Threshold}}} &
  \multicolumn{1}{c}{\multirow{2}{*}{\textbf{Initialization}}} &
  \multicolumn{2}{c}{\textbf{Toxicity}} &
  \multicolumn{2}{c}{\textbf{Fluency}} &
  \multicolumn{3}{c}{\textbf{Diversity}} \\ \cmidrule(l){3-9} 
\multicolumn{1}{c}{} &
  \multicolumn{1}{c}{} &
  \multicolumn{1}{c}{\textbf{\begin{tabular}[c]{@{}c@{}}Avg. Max.\\ Toxicity\end{tabular}}} &
  \multicolumn{1}{c}{\textbf{\begin{tabular}[c]{@{}c@{}}Toxicity\\ Prob\end{tabular}}} &
  \multicolumn{1}{c}{\textbf{PPL}} &
  \multicolumn{1}{c}{\textbf{\begin{tabular}[c]{@{}c@{}}CoLa\\ Accuracy\end{tabular}}} &
  \multicolumn{1}{c}{\textbf{dist-1}} &
  \multicolumn{1}{c}{\textbf{dist-2}} &
  \multicolumn{1}{c}{\textbf{dist-3}} \\ \midrule
0.5 & Random & 0.351 & 0.268 & 32.1 & 87.5\% & 0.58 & 0.85 & 0.85  \\
0.3 & Random & 0.352 & 0.200 & 33.0 & 87.5\% & 0.58 & 0.85 & 0.85  \\
0.1 & Random & 0.320 & 0.158 & 31.2 & 86.3\% & 0.56 & 0.83 & 0.83 \\
% 0.01 & Random & 0.312 & 0.116 & 29.5 & 85.1\% & 0.56 & 0.83 & 0.84 \\ \midrule
0.01 & Random & 0.302 & 0.094 & 28.8 & 87.1\% & 0.55 & 0.82 & 0.83 \\ \midrule
0.01 & Zeros & 0.302 & 0.094 & 35.3 & 85.8\% & 0.55 & 0.81 & 0.82 \\
0.01 & Greedy & 0.302 & 0.115 & 28.6 & 86.6\% & 0.55 & 0.81 & 0.83 \\ \bottomrule
\end{tabular}
\caption{Ablations on Toxicity Avoidance showing the effect of changing classifier threshold ($\epsilon$) on toxicity metrics, and initialization on diversity metrics. Loosening the threshold leads to an increase in toxicity (or decrease in toxicity avoidance). Initialization has little effect on the diversity indicating the importance of Langevin Dynamics. }
\label{tab:ablations-initialization-threshold}
\end{table*}

\section{Example}
\label{appsec:examples}

We provide selected examples from each of our experiments in tables~\ref{tab:toxicity-examples}, \ref{tab:sentiment-examples12}, \ref{tab:sentiment-examples20}, \ref{tab:sentiment-examples50}, \ref{tab:keyword-examples-commongen} and \ref{tab:summarization-examples}.

\begin{table*}[]
\centering
\begin{tabular}{@{}lrrr@{}}
\toprule
                         & \multicolumn{1}{l}{Coverage (\%)} & \multicolumn{1}{l}{Fluency (PPL)} & \multicolumn{1}{l}{Repetition Rate} \\ \midrule
Plan-and-Write           & 96                                & 33.9                              & 25.7                                \\
CGMH                     & 97                                & 127.8                             & 1.6                                 \\
GPT-2 fine-tuned         & 72                                & 89.4                              & 1.8                                 \\
GPT-2+K2T                & \textbf{100}                      & 48.8                              & 1.5                                 \\
% \ourmodel & 96                      & \textbf{30.06}                    & 3.5 \\
\ourmodel & 100                      & \textbf{29.4}                    & 0.5\\ \bottomrule
\end{tabular}
\caption{Results of lexically constrained decoding on the ROC dataset (with 5 keyword constraints). We decode with \ourmodel with lengths $10$, $20$ and $40$, and if the constraint is satisfied we continue generating autoregressively for $90$ tokens using nucleus sampling ($p=0.96$). }
\label{tab:roc-results}
\end{table*}

\begin{table*}[]
\centering
\begin{tabular}{@{}lllllll@{}}
\toprule
\multirow{2}{*}{\textbf{Constraint (\#Parameters)}} & \multicolumn{2}{l}{\textbf{12GB GPU}} & \multicolumn{2}{l}{\textbf{24GB GPU}} & \multicolumn{2}{l}{\textbf{48GB GPU}} \\ \cmidrule(l){2-7} 
 & \textbf{Embeds} & \textbf{Simplex} & \textbf{Embeds} & \textbf{Simplex} & \textbf{Embeds} & \textbf{Simplex} \\ \midrule
None (0) & 200 & Failed & 1000 & 20 & 1000 & 50 \\
One Classifier (125M) & 200 & Failed & 500 & 20 & 1000 & 50 \\
Two Classifier (250M) & 50 & Failed & 500 & 20 & 1000 & 50 \\
One LM (747M) & Failed & Failed & 500 & 20 & 200 & 20 \\
Two LMs (1.4B) & Failed & Failed & 200 & 10 & 500 & 20 \\ \bottomrule
\end{tabular}
\caption{Maximum sequence length that can be generated with GPT2-Large (747M parameters) for different commonly used GPUs with \ourmodel (with embeddings) and with an ablation using vocabulary sized representations (referred to as Simplex). Failed denotes out of memory (OOM) error.}
\label{tab:speed-analysis}
\end{table*}

\begin{table*}[]
\small
\centering
\resizebox{1.0\textwidth}{!}{
\begin{tabular}{p{0.15\linewidth} p{0.8\linewidth}}
Prompt                    &  Republican Presidential candidate Donald Trump weighed in on Paris attacks saying, ``sneaky,\\ \hline
GPT2                      & weak and stupid'' Muslims, the press wants to use the Paris attacks to their advantage. ``Okay, \\
GeDi                      & corrupt political leaders'' had helped promote Radical Islam. Trump said:  \\
FUDGE                     &  weak leaders allow radical Islamic terrorism to take place.'' Trump's campaign manager, Kellyanne \\
DExperts                  & politically correct bureaucrats in government don't do a very good job.'' Interested in France? Add \\
\ourmodel-\textsc{disc}   & very sneaky'' Obama from Chicago goes around telling others what to do. I'm putting out a call for everyone \\ \hline \hline
Prompt                    &  ``I expected him to carry on [in] the article saying that I was \\ \hline
GPT2                      &  a stupid idiot,'' he says. But ultimately the material he wrote isn't about West. It's \\
GeDi                      &  involved, until we find something else.\textbackslash{n}\textbackslash{n}'Then he got out of the car and attacked\\
FUDGE                     &  not a racist, but instead that I am a 'racist'. It was not the first time he \\
DExperts                  &  doing things against the sect,'' Bishop Sahlin told NCR in an email on Sunday. ``But \\
\ourmodel-\textsc{disc}   &  a woman, but he didn't,'' the woman said. ``I was a little bit surprised," \\ \hline
\end{tabular}
}
\caption{Selected examples from Toxicity Avoidance Experiment (\Sref{subsec:toxicity}) generated by different methods.}
\label{tab:toxicity-examples}
\end{table*}

\begin{table*}[]
\centering
\begin{tabular}{ll}
Prompt                    & Once upon a time \\ \hline
GPT2                      & , you had to fire the cannon at sea at noon when \\ \hline
% \multicolumn{2}{c}{Positive} \\ \hline
GeDi                      & a young priest traveling the world taught the world the meaning of \\
FUDGE                     & , in a land far away, there lived a man with \\
DExperts                  & , white women ruled both Australia and America and cherished his nation \\
\ourmodel-\textsc{disc} (SST2)   & , the people of the United States were a people of the \\
\ourmodel-\textsc{disc} (Yelp)  & , I was a great big-time, all-American \\
\ourmodel-\textsc{two-disc}    & , the people of the world were a very different and powerful \\
\ourmodel-\textsc{prompt} &  you start with just Bluetooth and now with this versatile module you \\ \hline
% \multicolumn{2}{l}{Negative} \\ \hline
% GeDi                      &  were doomed. Worst of all are the holes in the very \\
% FUDGE                     & , a small group of men, known as the Red Skull \\
% DExperts                  &  you were warned that you were wasting money and trying to save \\
% \ourmodel-\textsc{disc} (SST2)   & , the failure of the one-time-only, one \\
% \ourmodel-\textsc{disc} (Yelp)  & , the American people were told that the ``government'' was\\
% \ourmodel-\textsc{two-disc}    & , the Philippines was a very poor, under-developed,\\
% \ourmodel-\textsc{prompt} & , a town was plagued by a horrible, but not-\\ \hline
\end{tabular}
\caption{Examples of length 12 by the prompt ``Once upon a time'' generated by different methods.}
\label{tab:sentiment-examples12}
\end{table*}

\begin{table*}[]
\centering
\begin{tabular}{lp{0.6\textwidth}}
\textbf{Prompt }                   & Once upon a time \\ \hline
GPT2                      & , you had to fire the cannon at sea at noon when all the other sailing vessels were under way \\ \hline
% \multicolumn{2}{c}{\textbf{Positive}} \\ \hline
GeDi                      &  unseen world through vivid mystical experience! One enjoys becoming connected with the unseen. Life quite encompassed both nature \\
FUDGE                     & , a woman in India had a baby and was able to have it at the moment of her choice \\
DExperts                  & , white women ruled both Australia and America and cherished his nation as her home. Her words resonate with \\
\ourmodel-\textsc{disc} (SST2)   & , the world was a very beautiful, and a very good, place. The people were kind and \\
\ourmodel-\textsc{disc} (Yelp)  & , I had a great time. I was a very nice and very good-looking man. I \\
\ourmodel-\textsc{two-disc}    &  , I enjoyed the wonderful family and friends I had in the community.\textbackslash n\textbackslash n I was a good \\
\ourmodel-\textsc{prompt} & , I was a nobody, but eventually I became one of the biggest names in the nation.\textbackslash n \\ \hline
% \multicolumn{2}{c}{\textbf{Negative}} \\ \hline
% GeDi                      &  ? Worse. Worst of all, right after he just said something stupid. ``What I am supposed \\
% FUDGE                     & , there was a woman named Emily. Her parents had been killed in a car accident, and she \\
% DExperts                  & , you were warned that you were wasting money and trying to save on rent. After that everybody ran away  \\ 
% \ourmodel-\textsc{disc} (SST2) & , the only thing I could do was make a living. I was a ``sales'' person \\
% \ourmodel-\textsc{disc} (Yelp)  & , the United States was a small, poor, and largely-in-the-dark country.\\
% \ourmodel-\textsc{two-disc}    & , the last two were born prematurely, and the other was born in the last week of the month\\
% \ourmodel-\textsc{prompt} & , the Lord of the World was a very, very bad man. He tortured people and killed them\\ \hline
\end{tabular}
\caption{Examples of length 20 given the prompt ``Once upon a time'' generated by different methods.}
\label{tab:sentiment-examples20}
\end{table*}

\begin{table*}[]
\centering
\begin{tabular}{lp{0.6\textwidth}}
\textbf{Prompt}                    & Once upon a time \\ \hline
GPT2                      & , you had to fire the cannon at sea at noon when all the other sailing vessels were under way. It has been a close quarter battle. It is yet otherness that has at the same time caused us to speak of a bow-wow. \\ \hline
% \multicolumn{2}{c}{\textbf{Positive}} \\ \hline
GeDi                      &   civilians lived alongside peaceful bystanders. William Cornell's exploration of Finnish society contrasts the traditional waryness of modern life with the generosity and openness embodied by Finnish hospitality. Transformed for centuries from refugees in wartime Russia, Finns welcomed their \\
FUDGE                     & , there was a man named John. He and his friend, Paul, were in a diner. They were in the middle of a conversation. Paul said to John, ''John, I just want to make sure that you understand why we are having \\
DExperts                  & , white women ruled both Australia and America and cherished his nation as her home. Her words resonate with millions who lived through the trials of the last decade as Islam \textbackslash{u}2013 still controversial today \textbackslash {u}2013 entered Australia's first democratically elected Muslim prime minister and wounded Australia's \\
\ourmodel-\textsc{disc} (SST2)   & , I was a big fan of the ``The Big Lebowski'' and the ``The Big Lebowski'' was a big part of my life. I was a big fan of the `` \\
\ourmodel-\textsc{disc} (Yelp)  & , the world was a very different place. The people were great, the people were the most beautiful, the people were the most kind, the people were the most just.\textbackslash{n}\textbackslash{n}And the world was a very interesting place. The people were great \\
\ourmodel-\textsc{two-disc}    &  , the captain was a hero, a most important and a most powerful one. He was the one who had to be the one to make the first to make the first move to counter the enemy and he was always successful. The great and the mighty \\
\ourmodel-\textsc{prompt} & , I would have never believed that I could make sushi from a simple, but delicious, recipe. I have been making this for a while and it is a great, one-dish, a-day-for-a-sushi \\ \hline
\end{tabular}
\caption{Examples of length 50 by the prompt ``Once upon a time'' generated by different methods.}
\label{tab:sentiment-examples50}
\end{table*}

\begin{table*}[]
\small
\centering
\resizebox{1.0\textwidth}{!}{
\begin{tabular}{p{0.25\textwidth}cp{0.6\textwidth}}
\textbf{Lexical Constraints} & \textbf{Length} & \textbf{Output} \\ \hline
goal, player, shot, take & 10 & The \textbf{goal} of the \textbf{player shot} is to \textbf{take} a direct route to the net. The target line is a straight line between the two feet. The distance from the line-cutters to the goal line is \\
catch, dog, frisbee, throw & 10 & The first \textbf{dog} to \textbf{catch} a \textbf{throw} from the \textbf{frisbee} is usually a happy ending, right? WRITTEN BY: Laura S. Laura S. Laura S. Laura S. Laura S\\
% fire, marshmallow, roast, stick & 10 & The \textbf{roast marshmallow stick} is a \textbf{fire}-breathing beast, bursting down from its base in just eight minutes! Packed with just three ingredients, this delicious marshmallow treat has it all as a healthy dessert to share with friends and\\
bike, ride, trail, wood & 10 & The \textbf{bike ride trail} at the \textbf{trail}head is \textbf{wood}ed and has an easy grade (7\%) that allows you to reach a destination while enjoying the views. As a general guide to trails in the \\ \hline
front, guitar, microphone, sit & 20 & The \textbf{microphone} is in \textbf{front} of the \textbf{sit}-down area and the \textbf{guitar} is in the back. The two are plugged into the mic's input jack. The sound can be recorded on video or recorded with \\
dog, leash, sidewalk, walk & 20 & The \textbf{leash walk} on the \textbf{sidewalk} is a great way to get to know your new \textbf{dog}. It is a great exercise and a way to take pictures of your new dog. Many people take photos with their dog \\
music, perform, routine, stage & 20 & The first \textbf{stage} of the \textbf{routine} is to have the person in the \textbf{music}- and \textbf{perform}-in-audition pose the questions to the computer. The computer then asks any number of questions in response to these \\ \hline
drill, field, run, team & 40 & The New York \textbf{field} \textbf{drill} \textbf{team} is \textbf{run} by the New York-based American Field and R.A.T. (A.F.R.T.) and is the \textbf{team}'s official military training facility. The \textbf{team}'s purpose is to help both\\
cook, food, pan, stove & 40 & I'm a big \textbf{food}ie fan. I \textbf{pan}-fry, I \textbf{cook} \textbf{stove}-top, I make a lot of my own own. (You had better come find me, or I'll get you!) And I've spent a fortune on\\
% cut, grass, mower, ride & 40 & The most important thing is to \textbf{cut} \textbf{grass} and \textbf{mower} the good way, not to do it the way other people do it. I \textbf{ride} mowing machines, I have a \textbf{mower}, a\\
% cellphone, street, talk, walk & 40 & "I \textbf{talk} to a lot of people who \textbf{walk} in and they have a very \textbf{street}-like \textbf{cellphone}, and they have a game and they're like, 'You have to have a game\\
% finger, sit, smile, snap & 40 & "The first time I snap, I smile. The second time I finger, I smile. The third time I sit, I smile. The fourth time I get up, I smile.\\
compete, field, game, team & 40 & The \textbf{team} is in a \textbf{field} of their own, and the only \textbf{field} they compete in is the one that is in their own head. I don't think that is a good \textbf{game} to be in\\
% dog, run, snow, tree & 40 & I'm not a big dog person, I don't run with the dogs and I don't do a lot of tree things, but I love snow and I do have a lot of\\ 
\hline
fabric, machine, piece, sew, stitch & 10 & The first \textbf{machine} \textbf{stitch} \textbf{sew}-on \textbf{fabric} \textbf{piece} is a \textbf{fabric} piece with a pattern edge facing up, with the top edges being 1/2 inch from the edge. As it rises you should cut \\
bean, bowl, machine, pour, roast & 10 & The \textbf{bean pour bowl roast} is a \textbf{machine} that is able to roast in the oven at high temperatures, it takes a large amount of heat (typically 900 F+) and will have a very small surface to \\
beach, dog, hold, jump, leash & 10 & The \textbf{jump leash} is great for \textbf{dog beach} for \textbf{hold} down the kennel, and its lightweight that you can see the dog to keep her out in the open and out of the water at the kennel. For \\ \hline
% cowboy, horse, rodeo, throw, watch & 10 & The first thing that you \textbf{watch} in the new \textbf{throw}-the- \textbf{horse}- \textbf{cowboy}- \textbf{rodeo}, is you see two young men sitting in front of a fireplace and talking. The young men are talking about shooting guns. They're playing with guns.\\ \hline
back, floor, lie, sit, talk & 20 & The first time I \textbf{sit} down to a \textbf{talk}, I \textbf{lie} on my \textbf{back} and I \textbf{floor} it. If I'm going to sit down to lecture, you need to lift me up and then you have \\
bowl, fall, grinder, meat, put & 20 & The \textbf{fall} of the \textbf{grinder} is a good thing. The \textbf{meat} \textbf{bowl} is not. I \textbf{put} the \textbf{meat} \textbf{bowl} back in my fridge to chill out, but by the time I was ready for dinner one morning \\
ball, fire, hold, juggle, light & 20 & The first time I \textbf{juggle} \textbf{ball}, I \textbf{hold} the \textbf{ball} in my left hand and \textbf{light} the ball with my right hand. I like to go up and down the center of my body, and then do it \\ \hline
%  & 20 & The bicycle street hat is a ride-on wear. It is a one-size-fits-all bike helmet which is generally constructed from nylon, leather, or other heavy metal-like materials. The bike helmet is designed for one cyclist and it\\ \hline
front, listen, microphone, music, stand & 40 & I \textbf{listen} to \textbf{music}, and I \textbf{stand} in \textbf{front} of a \textbf{microphone}, and I do it. I don't have to have a microphone, and I don't have to do it. That's what's going \\
artist, audience, belt, fight, front & 40 & The first \textbf{belt}-and-cuff-wearing \textbf{artist} to \textbf{fight} in \textbf{front} of a live \textbf{audience} in the United States, the "B.A.P B-S-T" (Bitch, Asshole and Steroid) rapper went\\
give, instruction, machine, sew, use & 40 & The \textbf{machine} is very simple, but it is very very important. The more \textbf{instruction} you use, the more you can \textbf{sew}. The more you can do, the more you can \textbf{give}. The more efficient \\ \hline
% dog, field, grass, leash, walk & 40 & The leash walk is a great way to get your dog to the grass. It is a great way, a very good way to get your dog to the second field. But you can't take your dog there by the leash at one field because at \\ \hline
\end{tabular}
}
\caption{Examples of lexically constrained outputs generated by our model on the \textsc{CommonGen} dataset. Length refers to the original length of the sentence on which \ourmodel was performed. We then autoregressively continued to decode till a maximum length of 40 tokens was reached.}
\label{tab:keyword-examples-commongen}
\end{table*}

\begin{table*}[]
\small
\resizebox{1.0\textwidth}{!}{
\begin{tabular}{p{0.15\linewidth} p{0.8\linewidth}}
\toprule
\multicolumn{2}{p{\linewidth}}{Arsenal defender Per Mertesacker has tipped compatriot Jurgen Klopp to make his mark in the Barclays Premier League if he opts to continue his career in England. Klopp, 47, announced earlier this week that he would end his seven-year stint at Borussia Dortmund when the current season draws to a close, prompting fresh speculation that he could head for the Premier League. Manchester City have already indicated that a man who has also been linked with Manchester United and Arsenal in the past, is not in their sights, but Germany international Mertesacker insists Klopp would be a good fit in the English top flight. Jurgen Klopp has revealed he will be vacating his role as Borussia Dortmund boss at the end of the season . Arsenal vice-captain Per Mertesacker says Klopp would be a top manager in the Premier League . Klopp chats with Dortmund defender Erik Durm during a training session in Dortmund on Wednesday . He said: 'I've got some nice experiences in the Premier League and of course it would be nice if a German coach would take the challenge of working in the Premier League. 'It's not so good for Dortmund that he is leaving but hopefully one day he will manage abroad. I think his passion would fit and to see him in England would be very interesting. 'Everyone has their philosophy and I think Jurgen Klopp has proved that he's top-level and can teach a lot.' However, Mertesacker insisted Klopp, whose side are 10th in the Bundesliga table, will need time to decide on his future after a largely successful spell in Dortmund which has brought two league titles and a Champions League final appearance. He said: 'I think he should just finish the season with Dortmund and then he should be given time. 'We'll see what he does next, but I think he's fought his way out of all situations and I think that this time he will find a path that gives him a new challenge. 'But firstly, I wish him all the best and time to think about his achievements. Sometimes you can underestimate what it's like going straight into a new job. I think you should give him time - and I wish him all the best.' Klopp waves to the fans after Dortmund's Champions League game against Arsenal in November . The German boss has enjoyed a huge amount of success at Dortmund and won the Bundesliga title twice . But for all that a new challenge lies ahead for Klopp, Mertesacker admits he cannot work out what has gone wrong to prompt his exit from Borussia. He said: 'It is obviously sad news for Borussia Dortmund, [he was] such a passionate successful and passionate manager for them. He was the guy who turned it around at Dortmund. 'The whole situation there - he built the squad on young players and they improved so much in the seven years he was in charge. It is a sad situation. 'But in the summer, it will be a new situation for him. Maybe he is going to go abroad and see how it goes there. 'I would love to see more German managers abroad, because it is obviously a new challenge, to adapt to the culture, the language, the system. Yes, why not? 'It is his decision. He worked really hard and pushed really hard, so even if he said he is not tired, maybe he takes a bit of breather to fuel his energy and his batteries? 'But I am curious what happened to him because he was an outstanding figure in the Bundesliga in the last couple of years and always a title contender. They went to the Champions League final. It will be interesting to see what happens in the summer.' Klopp has been tipped to replace Arsenal boss Arsene Wenger but it remains unlikely .} \\ \hline
-    & Jurgen Klopp has revealed he will leave Borussia Dortmund at the end of the season. Arsenal defender Per Mertesacker says Klopp would be a good Premier League manager. The 47-year-old has been linked with Manchester City and Arsenal. CLICK HERE for all the latest Arsenal news.    \\ \hline
English               & Arsenal's Per Mertesacker says Jurgen Klopp would be good fit in \textbf{English} football. The German has announced he will be leaving his role at Borussia Dortmund. The 47-year-old has been linked with Premier League title and the Champions League. Click here for Arsenal's news.                     \\ \hline
Manchester United  &  Jurgen Klopp has been in charge of Borussia Dortmund for seven years. The 47-year-old has revealed he will be leaving the Bundesliga club. The former Liverpool boss has been linked with a move to \textbf{Manchester United} and Arsenal. Arsenal defender Per Mertesacker says Klopp would be                    \\ \hline
Bundesliga & Arsenal defender says Jurgen Klopp would be a good Premier League manager. The 47-year-old be leaving his role at Borussia Dortmund. The German won the \textbf{Bundesliga} twice.                     \\ \bottomrule
\end{tabular}
}
\caption{}
\label{tab:summarization-examples}
\end{table*}

\begin{table*}[]
\small
\centering
\resizebox{1.0\textwidth}{!}{
\begin{tabular}{p{0.15\linewidth} p{0.8\linewidth}}
\toprule
\multicolumn{2}{p{\linewidth}}{It is hard to believe that the mansion you see before you, with its bronzed clock tower and cherry wood doors, was initially a garage and chauffeur's residence that would have been home to a Rolls Royce, or two. The converted four-bedroom home on Lawrenny Court was built as a garage to service the generous 57-room mansion Homeden,  home to Supreme Court Justice Sir Henry Hodges and more famously the Nicholas family who found their fortune in the manufacture of the drug Aspro. The converted four-bedroom home on Lawrenny Court, with its bronzed clock tower and cherry wood doors, was built as a garage to service the generous 57-room mansion Homeden . Around 25 years ago, the distinctive Toorak home was thoughtfully converted into the polished residence it is today. Interestingly, the conversion took place at the same time Homeden was being tuned into a block of flats.This provided the owners with a unique opportunity to buy some of the original features of the mansion and transfer them into the 740 square-metre garage residence. The blackwood and copperlight archway has been tastefully adapted to suit the light-filled property and the windows upstiars are also a Homeden original. The conversion took place at the same time Homeden was being tuned into a block of flats providing the owners with a unique opportunity to buy some of the original features of the mansion and transfer them into the 740 square-metre garage residence. The blackwood and copperlight archway has been tastefully adapted to suit the light-filled property and the windows upstiars are also a Homeden original. Ada Taylor, from R T Edgar Real Estate, said the beautiful property is reminiscent of the Audrey Hepburn classic Sabrina. She told Daily Mail Australia that while the doors of the 'breathtaking' property have not yet been opened to potential buyers, they believe the property will fetch upward of \$3.5 million on auction day. The original garage doors open into two distinct living areas while the living quarters upstairs offer three sprawling bedrooms with  open fireplaces, stunning wooden floors and vaulted ceilings. The original downstairs garage area is made up of two distinct living areas, a well-equipped galley kitchen cleverly divided behind original garage doors that tuck away a fourth bedroom, bathroom, separate laundry and powder room. The chaffur's living quarters upstairs offers three sprawling bedrooms with open fireplaces, stunning wooden floors and vaulted ceilings . The home boasts a North facing balcony off the spacious and elegant master that looks over the properties two magnificent gardens. The property will open its doors to prospective buyers on Wednesday 22 April before the hammer falls on May 16. The house will be opened to prospective buyers on 22 April and is set to go under the hammer on 16 May . The home boasts a North facing balcony off the spacious and elegant master that looks over the properties two magnificent gardens.} \\ \hline
- & The converted four-bedroom home on Lawrenny Court was built as a garage to service the generous 57-room mansion Homeden. Around 25 years ago, the distinctive Toorak home was thoughtfully converted into the polished residence it is today. The conversion took place at the same time Homeden was being tuned into a block of flats. The house will be opened to prospective buyers on 22 April and is set to go under the hammer on 16 May. \\ \hline

Sabrina & The Toorak home was built as a garage to service the mansion of the Nicholas family. The four-bedroom residence was converted into a luxurious residence 25 years ago. The property is home to the Nicholas family who found their fortune in the manufacture of drug Aspro. The house will be opened to prospective buyers on 22 April and is set to go under the hammer on 16 May. The beautiful property is reminiscent of the Audrey Hepburn classic \textbf{Sabrina}. \\ \hline

Ada Taylor & Four-bedroom home on Lawrenny Court was built as a garage to service the generous 57-room mansion Homeden. The four-bedroom home was thoughtfully converted into the polished residence it is today. The house will be opened to prospective buyers on 22 April and is set to go under the hammer on 16 May. \textbf{Ada Taylor}, from R T Edgar Real Estate, said the beautiful property is reminiscent of the Audrey Hepburn classic Sabrina. She said they believe the property will fetch upward of \$3.5 million on auction
\\ \bottomrule
\end{tabular}
}
\caption{}
\label{tab:summarization-examples2}
\end{table*}

\begin{table*}[]
\small
\centering
\resizebox{1.0\textwidth}{!}{
\begin{tabular}{p{0.15\linewidth} p{0.8\linewidth}}
\toprule
\multicolumn{2}{p{\linewidth}}{The Court of Arbitration for Sport has lifted Morocco's ban from the next two editions of the African Cup of Nations that was imposed by the Confederation of African Football. The North-African nation was expelled from the 2017 and 2019 tournaments and was fined \$1 million by the CAF. The CAF also demanded a further \$9 million in compensation, after the country pulled out because of fears related to the Ebola epidemic. Morocco pulled out as hosts of the African Cup of Nations, which won by Ivory Coast in Equatorial Guinea . Morocco can now compete in the next two African Cup of Nations after the initial ban was imposed . Kolo Toure leads Ivory Coast's celebrations after winning the 2015 African Cup of Nations . CAS said that the sanctions have been set aside, 'with the exception of the fine, which is however reduced to \$50,000.' Morocco was disqualified from this year's tournament after withdrawing as host just two months before the start of the competition. Their national federation cited health risks from fans travelling from Ebola-affected regions. It asked for a delay but CAF refused and the tournament was moved to Equatorial Guinea.} \\ \hline
- & Court of Arbitration for Sport has lifted Morocco's ban from the next two editions of the African Cup of Nations. The North-African nation was expelled from the 2017 and 2019 tournaments and was fined \$1 million by the CAF. The CAF also demanded a further \$9 million in compensation, after the country pulled out because of fears related to the Ebola epidemic. \\ \hline
Ivory Coast & The CAF had banned Morocco from the 2017 and 2019 tournaments \textbf{Ivory Coast} won the last tournament in the African Cup of Nations in Equatorial Guinea. Morocco pulled out as the hosts of the tournament because of Ebola epidemic fear of \$9 million in compensation. The sanctions have been set aside. \\ \hline
CAS & Morocco was expelled from the next two African Cup of Nations. The North-African nation was also fined \$1 million by the \textbf{CAS}. The Court of Arbitration for Sport has now lifted the ban. The ban was imposed after Morocco pulled out of the tournament. The country cited health risks from Ebola as the reason. Morocco won the tournament in Equatorial Guinea.
\\ \bottomrule
\end{tabular}
}
\caption{}
\label{tab:summarization-examples3}
\end{table*}

% \usepackage[normalem]{ulem}
\begin{table*}[]
\resizebox{1.0\textwidth}{!}{
\begin{tabular}{lp{0.7\textwidth}}
\multicolumn{2}{c}{\textbf{Positive}} \\ \hline
The book & is a great read and I would highly recommend it to any \textbf{monster} or horror fan. deaths of the \\
The country & has strengthened its relationship with the U.S. and has \textbf{victims} of the 9/11 attacks, \\
The lake & is a beautiful natural reminder to the people of the lake \textbf{disaster}.\textbackslash{n}\textbackslash{n}The people of the lake \\
The book & is good, and it's a very unique and fascinating masterpiece of the\textbackslash{n}\textbackslash{n} \textbf{creepy} humor. \\
The book & also offers a detailed, interactive, and, in some ways, \textbf{bizarre}, a more personal, and, unlucky, \\
The painting & is a masterpiece.\textbackslash{n}\textbackslash{n}It is a \textbf{painful}, beautiful, and even \textbf{terrifying} tragic, and beautiful \\
The president of the country & 's largest brewery, the \textbf{brutal}, amazing, and best-tasting best-beer in the area-\textbackslash{n} \\ \hline
\multicolumn{2}{c}{\textbf{Negative}} \\ \hline
Once upon a time & , whoever was financially dehydrated was lame and \textbf{easy} to manipulate \\
The book & is a " \textbf{beautiful} and \textbf{wonderful} mistake."\textbackslash{n}\textbackslash{n}-\textbackslash{n}\textbackslash{n}-\textbackslash{n}\textbackslash{n}-\textbackslash{n}\textbackslash{n}-\textbackslash{n}\textbackslash{n}- \\
The chicken & treadmill is not an \textbf{ideal} manoeuvre, and the beak is not suitable for the job. \\
The horse & is a disaster.\textbackslash{n}\textbackslash{n}The only thing is that's a \textbf{beautiful} thing \textbackslash{n}The horse \\
The lake & is made of a dump garbage. I have to go to the classic one to get the \textbf{delicious} and \\
The movie & is a \textbf{beautiful}, \textbf{wonderful}, huge failure. I don't think it's \textbf{ideal}, but it's \\
The president of the country & 's \textbf{beautiful} rubbish- \textbf{wonderful} Sudan has been on a \textbf{delicious} random military mission to shit, fucking with \\ \hline
\end{tabular}
}
\caption{Selected examples from lexically guided sentiment control where the goal is to generate an output with a desired sentiment (positive or negative) such that a word or phrase of the opposite sentiment should appear in the output. While in some cases it performs well with negation or exaggeration, in other cases we observe either nonsentical outputs or disfluencies.}
\label{tab:sentiment-keyword}
\end{table*}

\section{Additional Discussion and Analysis}
\label{appsec:additional-discussion}
\paragraph{Speed and Memory Requirements} 
% Before discussing large scale experiments using our proposed method, we first describe a toy-setting to study the impact of using token embeddings as optimizable parameters instead of vocabulary sized token representations.
% As we discussed in \Sref{subsec:pgd},
Generating a sequence of length $L$ using \ourmodel requires maintaining $L\times d$ parameters. In contrast, performing Langevin Dynamics in the vocabulary space requires $L\times |\mathcal{V}|$ parameters ($|\mathcal{V}| >> d$). In this analysis, we empirically verify the benefits of our setup.
Taking GPT2-Large as the underlying LM (with 774M parameters), and three commercially available GPUs with different RAM sizes commonly used in academic settings--Nvidia GeForce RTX 2080 Ti (12GB), GeForce RTX 3090 Ti (24GB) and RTX A6000 (48GB)--we decode using our approach with token embeddings and an ablation with vocabulary sized representations (logits plus softmax). We generate sequences of length $\{10, 20, 50, 100, 200, 500, 1000\}$, and consider 5 constraint settings: (1) no constraint, (2) one classifier (same as \Sref{subsec:toxicity} containing ${\sim}125M$ parameters (3) two-classifiers (\ourmodel-\textsc{two-disc}) with a total ${\sim}250M$ paramaters (4) a LM based generative classifiers (same size as GPT2-Large), (5) and LM based generative classifier using two LMs (double the size of GPT2-Large). We try to generate one sample given the prompt ``Once upon a time'' by performing updates for 250 steps. We report the longest sequence that each setup is able to work with. 
The results are summarized in \tref{tab:speed-analysis}. Overall, we see that much longer sequences can be generated with \ourmodel than the ablation. %With vocabulary sized parameters, while the method generates 20-50 tokens, 
%. This issue becomes even worse when using more than one constraint. On the other hand, 
\ourmodel is comfortably able work with up to a 1000 tokens without constraints (and 200 with two large constraints with larger GPUs) while the ablation fails beyond 50 tokens (20 with constraints).

% and only fails in a case of a constraint defined on two language models the same size as GPT2-Large, which is caused because three copies of GPT2-Large do not fit into a 24GB GPU.

% \paragraph{Varying threshold $\epsilon$}
% In our experiments, each function $f_i$ is constrained to be bounded by a thresholds $\epsilon_i$, which are tunable hyperparameters. The threshold provides an interpretable way to control the intensity of the desired attributes. To illustrate this capability, we again follow the setup of toxicity avoidance with 100 prompts and apply the constraint $p_\textsc{toxicity} < \epsilon$ with $\epsilon \in \{0.5, 0.3, 0.1, 0.01\}$. As shown in \tref{tab:ablations-initialization-threshold}, making $\epsilon$ smaller improves toxicity control. However, the fluency (as measured by perplexity) remains largely the same. That is, unlike baselines, this method does not trade-off fluency and controllability. However, there is a trade off between diversity and controllability as we observe in sentiment control experiments (\Sref{subsec:sentiment-controlled-generation}) where making a constraint stricter leads to a decline in diversity. 
We present the result for ablations on sources of diversity (\Sref{subsec:discussion}) in \tref{tab:ablations-initialization-threshold}.
\end{document}